\documentclass[10pt,twocolumn,letterpaper]{article}

\usepackage{cvpr}              % To produce the CAMERA-READY version
\definecolor{cvprblue}{rgb}{0.21,0.49,0.74}
\usepackage[pagebackref,breaklinks,colorlinks,allcolors=cvprblue]{hyperref}

\usepackage[utf8]{inputenc} % allow utf-8 input
\usepackage[T1]{fontenc}    % use 8-bit T1 fonts
\usepackage{url}            % simple URL typesetting
\usepackage{booktabs}       % professional-quality tables
\usepackage{amsfonts}       % blackboard math symbols
\usepackage{nicefrac}       % compact symbols for 1/2, etc.
\usepackage{microtype}      % microtypography
\usepackage{xcolor}         % colors
\usepackage{amsmath}
\usepackage{amssymb}
\usepackage{mathtools}
\usepackage{amsthm}
\usepackage{multirow}
\usepackage[table]{xcolor}
\usepackage{color}
\definecolor{mygreen}{RGB}{146, 199, 113} 
\usepackage{graphicx}
\usepackage{adjustbox}
\usepackage{array}
\usepackage{booktabs}
\usepackage{dsfont}
\usepackage{makecell}
\usepackage{wrapfig}
\usepackage{algorithmic}
\usepackage{algorithm}

\def\paperID{2435} % *** Enter the Paper ID here
\def\confName{CVPR}
\def\confYear{2026}

\title{FINE: Factorizing Knowledge for Initialization of Variable-sized Diffusion Models}

\author{Yucheng Xie\textsuperscript{\rm 1,2}\; Fu Feng\textsuperscript{\rm 1,2}\; Ruixiao Shi\textsuperscript{\rm 1,2}\; Jianlu Shen\textsuperscript{\rm 1,2}\; Jing Wang\textsuperscript{\rm 1,2}\thanks{Corresponding authors}\; Yong Rui\textsuperscript{\rm 1,2}\; Xin Geng\textsuperscript{\rm 1,2}\footnotemark[1]\\
\textsuperscript{\rm 1}School of Computer Science and Engineering, Southeast University, Nanjing, China\\
\textsuperscript{\rm 2}Key Laboratory of New Generation Artificial Intelligence Technology and Its Interdisciplinary \\Applications (Southeast University), Ministry of Education, China\\
{\tt\small \{xieyc, fufeng, eric\_xiao, jlshen, wangjing91, xgeng\}@seu.edu.cn}\\
}

\begin{document}
\maketitle
\begin{abstract}
The training of diffusion models is computationally intensive, making effective pre-training essential. 
However, real-world deployments often demand models of variable sizes due to diverse memory and computational constraints, posing challenges when corresponding pre-trained versions are unavailable.
To address this, we propose FINE, a novel pre-training method whose resulting model can flexibly factorize its knowledge into fundamental components, termed learngenes, enabling direct initialization of models of various sizes and eliminating the need for repeated pre-training.
Rather than optimizing a conventional full-parameter model, FINE represents each layer’s weights as the product of $U_{\star}$, $\Sigma_{\star}^{(l)}$, and $V_{\star}^\top$, where $U_{\star}$ and $V_{\star}$ serve as size-agnostic learngenes shared across layers, while $\Sigma_{\star}^{(l)}$ remains layer-specific.
By jointly training these components, FINE forms a decomposable and transferable knowledge structure that allows efficient initialization through flexible recombination of learngenes, requiring only light retraining of $\Sigma_{\star}^{(l)}$ on limited data.
Extensive experiments demonstrate the efficiency of FINE, achieving state-of-the-art performance in initializing variable-sized models across diverse resource-constrained deployments. 
Furthermore, models initialized by FINE effectively adapt to diverse tasks, showcasing the task-agnostic versatility of learngenes.
\end{abstract} 

\section{Introduction}
\label{sec:intro}
Denoising diffusion models~\cite{ho2020denoising, austin2021structured, croitoru2023diffusion, guo2024diffusion} have recently emerged as a promising alternative to traditional Generative Adversarial Networks (GANs)~\cite{goodfellow2014generative, gui2021review}, due to their capacity to model highly complex data distributions. 
However, their substantial computational and memory requirements~\cite{wang2024patch, karras2024analyzing} have made training efficiency a key challenge in practice~\cite{hang2023efficient, zhang2024improving, xia2023diffir}.
Current approaches to improving the training efficiency of diffusion models, such as Parameter-Efficient Fine-Tuning (PEFT), enhance adaptability by injecting a small number of trainable parameters into frozen pre-trained backbones~\cite{qiu2023controlling, hulora, meng2024pissa, liudora, hyeonfedpara}.

\begin{figure}[t]
  \centering
  \includegraphics[width=\linewidth]{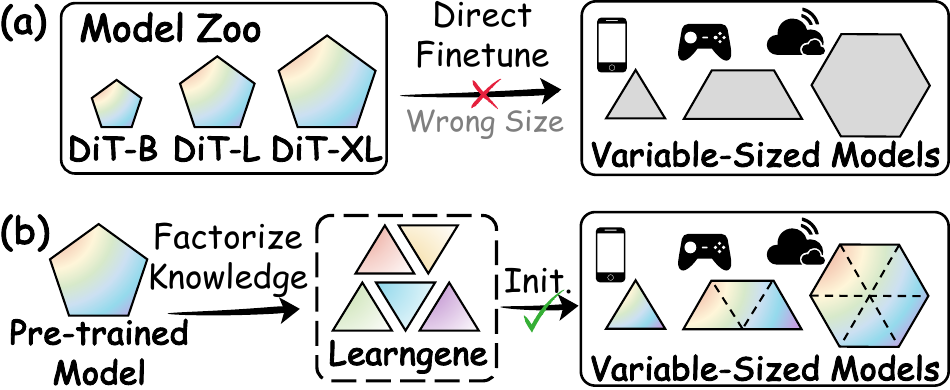}
  \vspace{-0.27in}
  \caption{(a) Pre-trained models are typically available only in standard model sizes, which may not meet specific requirements for deployment, thus necessitating time-consuming training from scratch. (b) We propose to pre-train models whose knowledge can be factorized into size-agnostic units, termed learngenes, enabling direct initialization of models with variable sizes as needed.}
  \label{fig:motivation}
  \vspace{-0.18in}
\end{figure}

While effective, the large number of parameters in pre-trained diffusion models limits their scalability and complicates deployment across heterogeneous hardware environments with varying computational and memory constraints~(Fig.~\ref{fig:motivation}a)~\cite{sheng2022larger, chen2019deep}. 
These limitations have motivated increasing interest in developing diffusion models with variable sizes to accommodate diverse deployment scenarios~\cite{lee2024multi, park2024denoising, ham2025diffusion}.
However, pre-trained diffusion models are typically released at a small number of fixed scales, making it impractical to pre-train and maintain models for all possible configurations. 
This limitation gives rise to a key question: \textit{Can we pre-train a unified model that can efficiently initialize diffusion models of varying sizes?}

% Model initialization plays a critical role in the convergence speed of neural networks~\cite{arpit2019initialize, huang2020improving}. 
Recently, the \textit{Learngene} framework, inspired by the natural transfer of genetic information, has emerged as a promising approach for leveraging pre-trained models to initialize variable-sized models~\cite{feng2023genes}. This framework integrates reusable, size-agnostic knowledge into compact units, termed ``learngenes'', which are then utilized to flexibly and efficiently initialize downstream models of varying sizes~\cite{xia2024transformer, feng2024wave, xie2024kind}.
Despite their promise, most existing learngene-based methods adopt heuristic, layer-specific strategies, where selected layers from a pre-trained model are manually reused to construct target models of different sizes~\cite{wang2023learngene, wang2022learngene, xia2024exploring}. 
However, such approaches overlook the intrinsic characteristics of diffusion-based image generation, where semantic consistency must be maintained across noise levels and layers. The layer-isolated design of existing learngene methods fails to capture such cross-layer dependencies, limiting their ability to model the hierarchical and temporally coupled representations essential to diffusion processes.
% As a result, their generality and adaptability across architectures remain limited, and their potential in generative domains has yet to be effectively explored.

To address these limitations, we propose FINE, an advanced pre-training framework within the \textit{Learngene} paradigm, in which the knowledge of the resulting model can be flexibly \textbf{F}actorized for \textbf{IN}itialization of diffusion models with variable siz\textbf{E}s.
To achieve this, FINE represents each layer’s weights as the product of $U_{\star}$, $\Sigma_{\star}^{(l)}$, and $V_{\star}^\top$, enabling scalable factorization in a formulation formally akin to SVD, rather than optimizing a conventional full-parameter model.
Unlike previous methods such as KIND~\cite{xie2024kind} and SVDiff~\cite{han2023svdiff}, which apply SVD independently to each layer, FINE introduces a \textbf{\textit{cross-layer weight sharing}} mechanism. 
Specifically, during pre-training, FINE jointly optimizes $U_{\star}$, $\Sigma_{\star}^{(l)}$, and $V_{\star}^\top$, where $U_{\star}$ and $V_{\star}$ are shared across layers as learngenes that encapsulate reusable, size-agnostic knowledge. 
The layer-specific matrices $\Sigma_{\star}^{(l)}$ are lightweight and adapt the shared representation to individual layers, enabling flexible initialization of models across variable-sizes.

% This cross layer weight sharing mechanism is implemented by training an auxiliary model, where each layer’s weight matrix is constructed as $W_{\star}^{(l)} = U_{\star} \Sigma_{\star}^{(l)} V_{\star}^\top$. The knowledge factorization is thus achieved implicitly through optimizing the auxiliary model under this shared parameterization.

As with other pre-training approaches, the computational overhead of FINE for pre-training a knowledge-factorizable model is a \textbf{\textit{one-time cost}}.
Initialization for new model sizes requires only training the lightweight $\Sigma_{\star}^{(l)}$, which operate in a compact parameter space~\cite{peng2024sam} and converge with significantly fewer steps (e.g., 0.3K vs. 300K for full pre-training).
We evaluate FINE mainly on image generation tasks using Diffusion Transformers (DiTs)~\cite{peebles2023scalable} as the backbone, and further demonstrate its generality by extending it to classification tasks with DeiT~\cite{touvron2021training}.
Experimental results demonstrate that FINE achieves state-of-the-art performance on variable-size initialization benchmarks, outperforming existing initialization and learngene-based methods.
Notably, FINE reduces FID by up to 4.89 (e.g., DiT-B $L_{10}$) and yields a $3n\times$ training speedup over full pre-training.
Furthermore, FINE generalizes well to new domains, achieving notable FID gains on CelebA ($\downarrow$0.28), LSUN-Bedroom ($\downarrow$2.60), and LSUN-Church ($\downarrow$2.01), demonstrating strong transferability.

Our main contributions are as follows: 
1)~We propose FINE, a novel pre-training method whose resulting model possesses flexibly factorizable knowledge, enabling efficient initialization of downstream models of variable sizes.
2)~We introduce the first comprehensive benchmark to evaluate the initialization capability of learngenes in image generation tasks.
3)~Extensive experiments validate the effectiveness of FINE, demonstrating state-of-the-art performance compared to other initialization and learngene methods.

\section{Related Work}
\paragraph{Efficient Training and Model Initialization}
Training diffusion models is computationally expensive, with growing demands on time and GPU resources becoming key bottlenecks~\cite{wang2024patch, karras2024analyzing}.
To improve training efficiency, most existing approaches rely on Parameter-Efficient Fine-Tuning (PEFT) methods~\cite{qiu2023controlling, hulora, meng2024pissa}. 
However, these methods heavily rely on pre-trained models and lack the flexibility to adapt to variable model sizes, making them unsuitable for deployment in heterogeneous hardware environments where no appropriately sized pre-trained models exist.
While other strategies~\cite{hang2023efficient, wang2024patch} aim to optimize diffusion model training, they remain constrained by specific assumptions.
% Several strategies have been proposed to optimize diffusion model training. For example, Min-SNR weighting~\cite{hang2023efficient} balances conflicting gradients across time steps, and Patch Diffusion~\cite{wang2024patch} reduces costs by training on image patches. While effective, these methods are often limited by specific conditions or assumptions.

Model initialization also plays a crucial role in training efficiency.
Beyond traditional methods like He-init~\cite{chen2021empirical},  methods such as GHN~\cite{knyazev2021parameter, knyazev2023canwescale} employ hypernetworks to predict parameters for diverse architectures, while LiGO~\cite{wang2023learning} leverages smaller pre-trained models as initialization.
However, these methods are largely restricted to classification tasks. Diffusion models, by contrast, involve iterative noise conditioning and cross-layer attention, which disrupt simple layer-wise correspondences, making direct scaling or parameter transfer suboptimal for preserving generative consistency.

% and efficient initialization for diffusion models in image generation remains largely unexplored.
% Addressing this gap is essential for advancing the practical use of diffusion models across various applications.

\vspace{-0.15in}
\paragraph{Learngene}
Inspired by biological evolution~\cite{bohacek2015molecular, waddington1942canalization}, the \textit{Learngene} framework~\cite{feng2023genes, feng2024transferring} establishes a principled paradigm for the initialization of variable sized networks, especially in the absence of pre-trained models.
Previous learngene-based methods, such as Heur-LG~\cite{wang2022learngene} and Auto-LG~\cite{wang2023learngene}, leverage heuristic and meta-learning strategies to identify transferable layers for specific tasks. 
However, existing learngene-based methods predominantly adopt layer-isolated designs that neglect cross-layer dependencies~\cite{xie2024kind, xie2025divcontrol, feng2025knowledge}, thereby limiting their capacity to capture the hierarchical and temporally coupled representations essential for the efficient initialization of variable-sized diffusion models in image generation.

Recently, WAVE~\cite{feng2024wave} proposes a constraint-based pre-training framework for scalable model initialization, enforcing structural constraints such as Kronecker products or Tucker decomposition~\cite{xie2026self} on weight matrices to promote structured knowledge integration.
Building upon this paradigm, FINE factorizes knowledge during pre-training into fundamental components shared across layers, which can be adaptively recombined according to target model sizes, thereby enhancing both flexibility and efficiency.

\section{Methods}
\subsection{Preliminary}
\paragraph{Latent Diffusion Models}
Latent diffusion models perform the diffusion process in a latent space for improved efficiency.
Given an image $x$, it is first encoded into a latent representation $z$ via an autoencoder $\mathcal{E}$, where $z = \mathcal{E}(x)$. 
The diffusion model is then trained to reconstruct $z$ through a denoising process by minimizing the following objective:
\begin{equation}
    \mathcal{L}=\mathbb{E}_{z,c,\varepsilon,t}[||\varepsilon-\varepsilon_{\theta}(z_t, c, t)||^{2}_{2}]
\label{eq:loss}
\end{equation}
where $\varepsilon_{\theta}$ is the noise prediction network with parameters $\theta$, which is trained to predict the noise $\varepsilon$ added to the latent variable $z_t$ at timestep $t$, conditioned on vector $c$. 

\vspace{-0.18in}
\paragraph{Diffusion Transformers (DiTs)}
DiTs represent an advanced transformer-based architecture for latent diffusion models, which use a decoder with $L$ stacked layers for noise prediction. Each layer consists of 
Multi-Head Self-Attention (MSA) for cross-patch integration and the Pointwise Feedforward (PFF) layers for within-patch processing.

In the MSA module, the model uses $H$ attention heads. For each head $A_i$, self-attention is performed with matrices $Q_i$, $K_i$, and $V_i \in \mathbb{R}^{T \times d}$, where the associated parameter matrices $W_\text{q}^i$, $W_\text{k}^i$, and $W_\text{v}^i \in \mathbb{R}^{D \times d}$ define the transformation. The output for the $i$-th attention head $A_i$ is given by:
\begin{equation}
    A_i = \text{softmax}(\frac{Q_i K_i^\top}{\sqrt{d}})V_i\,,\; A_i \in \mathbb{R}^{T \times d}
\end{equation}
The outputs from all $H$ attention heads are concatenated and projected through a weight matrix $W_{\text{o}}$:
\begin{equation}
    \text{MSA} = \text{concat}(A_1, A_2, ..., A_{H}) W_\text{o}\,,\; W_\text{o} \in \mathbb{R}^{Hd \times D}
\label{equ:msa}
\end{equation}
For computational efficiency, the attention heads’ parameters $W_\text{q}^i$, $W_\text{k}^i$, and $W_\text{v}^i \in \mathbb{R}^{D \times d}$ across all attention heads are combined into a larger matrix $W_{\text{qkv}} \in \mathbb{R}^{D \times 3Hd}$.

The PFF layer comprises two linear transformations, $W_{\text{in}} \in \mathbb{R}^{D \times D'}$ and $W_{\text{out}} \in \mathbb{R}^{D' \times D}$, with a GELU~\cite{hendrycks2016gaussian} activation function applied between them:
\begin{equation}
    \text{PFF}(x) = \text{GELU}(xW_{\text{in}} + b_1)W_{\text{out}} + b_2
\label{equ:mlp}
\end{equation}
where $b_{1}$ and $b_{2}$ are biases, and $D'$ denotes the hidden layer dimension.

Additionally, DiTs utilize an adaptive norm (i.e., adaLN) with a parameter matrix $W_{\text{adaLN}}$, which adjusts according to the embedding vectors of the time step $t$ and condition $c$:
\begin{equation}
    \text{Concat}(\alpha, \beta, \gamma) = \text{adaLN}(c, t)
\label{equ:adaLN}
\end{equation}
where $\alpha$ and $\gamma$ are dimension-wise scale parameters and $\beta$ is the shift parameters.

Thus, for a DiT with $L$ layers, the complete set of weight matrices is represented as $\theta = \{W_{\text{qkv}}^{(1\thicksim L)}, W_{\text{o}}^{(1\thicksim L)}, W_{\text{in}}^{(1\thicksim L)}, W_{\text{out}}^{(1\thicksim L)}, W_{\text{adaLN}}^{(1\thicksim L)}\}$\footnote{$W_{\text{qkv}}^{(1\thicksim L)}$ denotes the set $\{W_{\text{qkv}}^{(1)}, W_{\text{qkv}}^{(2)}, \dots, W_{\text{qkv}}^{(L)}\}$, with similar notation applied throughout the paper.}.

\vspace{-0.15in}
\paragraph{Size-agnostic Knowledge}
Transformer architectures are composed of stacked blocks with identical configurations, enabling the emergence of knowledge that is invariant to network depth—referred to as size-agnostic knowledge. 
Recent studies have progressively uncovered these patterns. Mimetic initialization~\cite{trockman2023mimetic} identifies diagonal patterns in $W_q W_k^\top$ and $W_v W_{proj}$ within each block of pre-trained ViTs. 
TLEG~\cite{xia2024transformer} further reveals linear correlations among block parameters, while ShareInit~\cite{lan2019albert} and MiniViT~\cite{zhang2022minivit} demonstrate that reusing specific blocks can retain performance with reduced model capacity.
However, these findings are primarily focused on ViTs. FINE expands this exploration to DiTs, where size-agnostic knowledge is represented as shared singular vectors across layers within weight matrices. 

\subsection{Pre-training Knowledge Factorizable Models}
\label{sec:KF}
\begin{figure*}[t]
  \centering
  \includegraphics[width=\linewidth]{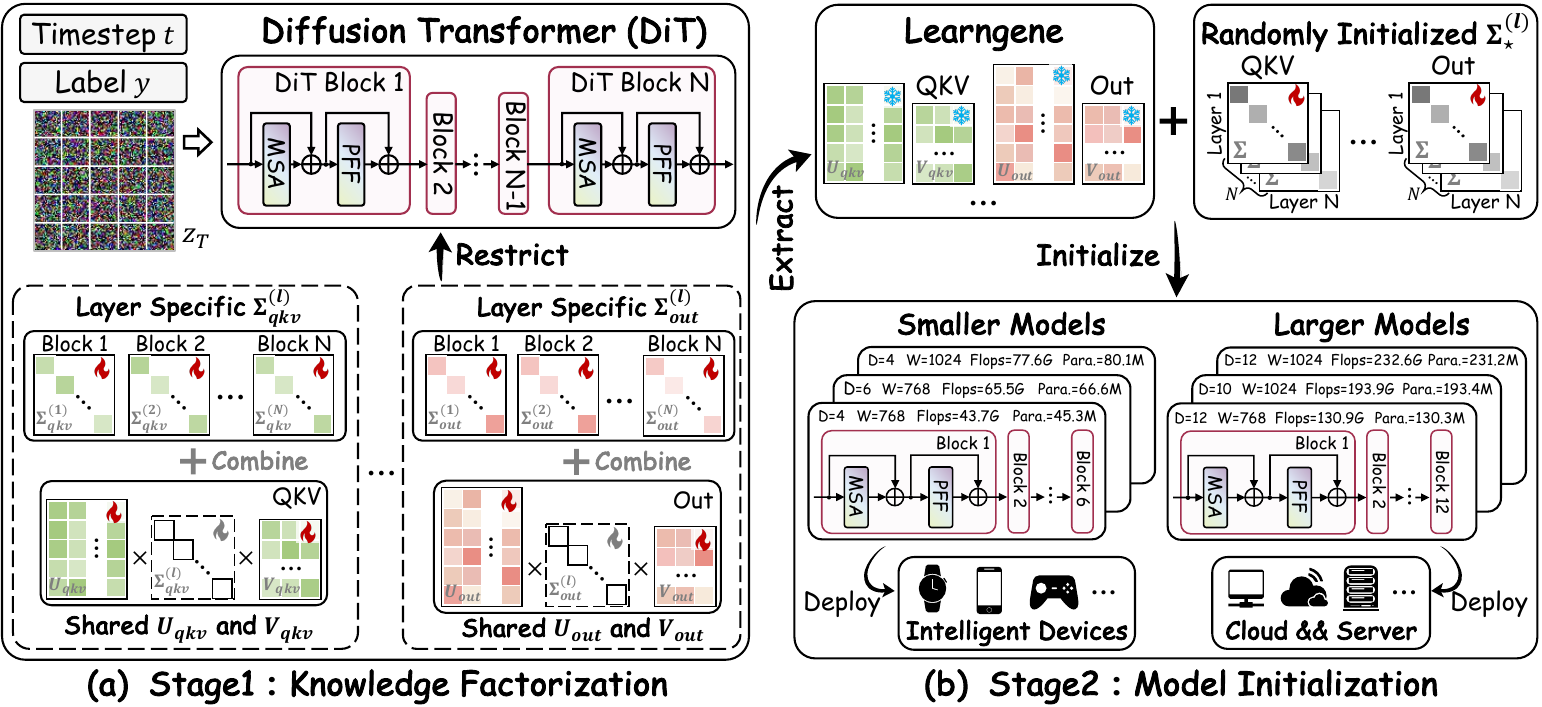}
  \vspace{-0.25in}
  \caption{Framework of FINE. (a) Knowledge within a Diffusion Transformer (DiT) is initially factorized into shared singular vectors, $U_{\star}$ and $V_{\star}$, and layer-specific singular values, $\Sigma_{\star}^{(l)}$, as described by Eq. (\ref{equ:svd}). 
  This factorization captures the shared, size-agnostic components of the model (i.e., learngenes), while $\Sigma_{\star}^{(l)}$ retains layer-specific variations. (b) During model initialization, only the singular values $\Sigma_{\star}^{(l)}$ need to be adapted based on the target model size. These values can be optimized using a small amount of data, while learngenes, represented by the shared $U_{\star}$ and $V_{\star}$, remain frozen.}
  \label{fig:form}
  \vspace{-0.1in}
\end{figure*}

To enable knowledge factorization during pre-training, we first decompose the weight matrices of each DiT block to better extract size-agnostic knowledge.
Recent studies~\cite{han2023svdiff, zhang2024spectrum, zhang2024spectral} have advanced SVD-based methods, primarily for parameter-efficient fine-tuning through layer-wise decomposition.
However, these methods overlook \textit{inter-layer shared knowledge}, leading to size-dependent deployment constraints and redundant storage due to uncoordinated layer-specific factorization.

In contrast, FINE emphasizes capturing size-agnostic knowledge shared across layers during factorization.
However, directly applying SVD to pre-trained models fail to enforce the sharing of $U_{\star}$ and $V_{\star}$ across layers, as SVD independently decomposes matrices without facilitating inter-layer knowledge sharing.
To address this, FINE adopts a reverse approach by first defining \textbf{\textit{shared singular vectors}} $U$ and $V$, alongside \textbf{\textit{layer-specific singular value}} $\Sigma^{(l)}$, to reconstruct weight matrices. 
These shared components, $U$ and $V$, serve as learngenes, encapsulating size-agnostic knowledge and enabling efficient initialization of models with variable sizes, as shown in Figure~\ref{fig:form}.

Given a DiT with $L$ layers and weight matrix types $\mathcal{T} = \{\text{qkv}, \text{o}, \text{in}, \text{out}, \text{adaLN}\}$, the weight parameters are denoted as $\theta = \{W_{\star}^{(l)}| \star\in \mathcal{T}, l\in [1, L]\}$.
We propose that, for each weight matrix type, the corresponding singular vectors $U_{\star}$ and $V_{\star}$ can be shared across all layers, thereby capturing size-agnostic knowledge.
The \textit{factorization rules} of each layer’s weight matrix is expressed as:
\begin{equation}
    W_{\star}^{(l)} \Leftarrow U_{\star}\Sigma_{\star}^{(l)} {V_{\star}}^\top
\label{equ:svd}
\end{equation}
Here, $\Leftarrow$ emphasizes that knowledge factorization is a reverse process, rather than directly applying SVD to pre-trained weight matrices.
$U_{\star}\in \mathbb{R}^{m_1 \times r}$ and $V_{\star}\in \mathbb{R}^{r \times m_2}$ are shared across layers of the same type (e.g., $W_{\text{qkv}}^{(1\thicksim L)}$ share the same $U_{\text{qkv}}$ and $V_{\text{qkv}}$), while $\Sigma_{\star}^{(l)}=\text{diag}(\boldsymbol{\sigma})$ is unique to each layer, with $\boldsymbol{\sigma}=[\sigma_1, \sigma_2, ..., \sigma_r]$. 

We further define the sets of shared singular vectors as $\mathcal{U}=\{U_{\star}| \star \in \mathcal{T}\}$, $\mathcal{V}=\{V_{\star}| \star \in \mathcal{T}\}$, and the set of singular values as $\mathcal{S}=\{\Sigma_{\star}^{(l)}| l\in [1, L], \star \in \mathcal{T}\}$.
To succinctly represent the construction rule, we abbreviate Eq.~\eqref{equ:svd} as:
\begin{equation}
    \theta = \mathcal{U}\mathcal{S}\mathcal{V}^\top
\label{equ:svd_sip}
\end{equation}
% To further reduce the number of trainable parameters in $\Sigma$, we apply parameters sharing within $\Sigma_{\star}^{(l)}$. 
% Specifically, if every $s$ parameters share the same value, the trainable parameters in $\Sigma_{\star}^{(l)}$ can be further reduced to $\frac{r}{s}$.
The pre-training of knowledge factorizable models is conducted under the constraint of Eq.~\eqref{equ:svd_sip}, formulated as the following optimization objective:
\begin{equation}
    {\underset{\mathcal{U}, \mathcal{S}, \mathcal{V}}{{\arg\min}} \, \mathcal{L}(\varepsilon_{\theta}(z_t, t, c), \varepsilon)},\;\;\text{s.t.}\, \theta = \mathcal{U}\mathcal{S}\mathcal{V}^\top
\label{equ:ob}
\end{equation}
where the loss $\mathcal{L}$ is defined in Eq.~\eqref{eq:loss}. 
Note that $\mathcal{L}$ is exclusively used to update the parameters of $\mathcal{U}$, $\mathcal{V}$ and $\mathcal{S}$ (Eq.~\eqref{equ:ob}), while the parameters of model $\theta$ are indirectly updated by being reconstructed under the rule in Eq.~\eqref{equ:svd} at each iteration. More details can be found in Algorithm~\ref{alg:algorithm}.

% This constraint enforces the sharing of $U_{\star}$ and $V_{\star}$ across layers, facilitating the capture of size-agnostic knowledge, while the layer-specific singular values $\Sigma_{\star}^{(l)}$ allow flexibility to adapt to variable model sizes.

% The auxiliary model is trained following standard diffusion model procedures, generating latent codes during denoising process to minimize 

% Upon completing the auxiliary model training, we successfully extract the final learngenes $\mathcal{G}$, which are formally represented as:
% \begin{equation}
%     \mathcal{G}=\{(U_{\star}, V_{\star})| \star \in \{\text{qkv}, \text{o}, \text{in}, \text{out}, \text{adaLN} \} \}
% \label{eq:G}
% \end{equation}

\subsection{Initialization of Variable-sized Models}
Current approaches primarily extract layer-specific learngenes~\cite{wang2022learngene, wang2023learngene, xia2024transformer}, which are manually stacked to initialize models of varying depths~\cite{xia2024exploring}. However, such heuristic designs introduce subjectivity and limit general applicability. This issue is particularly pronounced in diffusion models, where dynamically evolving inter-layer interactions during denoising make rigid stacking prone to coherence disruption.

Unlike prior methods, FINE enables \textbf{\textit{manual-free}} initialization by allowing learngenes to adaptively tailor to target model sizes in a data-driven manner, overcoming the limitations of prior approaches.
For initializing a target model with parameters $\theta_{\text{tgt}}$, the learngenes (i.e., shared $\mathcal{U}$ and $\mathcal{V}$) are frozen, while the layer-specific singular values $\mathcal{S}_{\text{tgt}}$ are adapted to the target model size and randomly initialized.
The objective is to achieve optimal initialization by optimizing these singular values:
\begin{equation}
    {\underset{\mathcal{S}_{\text{tgt}}}{{\arg\min}} \, \mathcal{L}(\varepsilon_{\theta_{\text{tgt}}}(z_t, t, c), \varepsilon)},\;\;\text{s.t.}\, \theta_{\text{tgt}} = \mathcal{U}\mathcal{S}_{\text{tgt}}\mathcal{V}^\top
\label{equ:ob2}
\end{equation}
Since $\mathcal{S}$ contains relatively few parameters, forming a compact parameter space~\cite{peng2024sam}, it can be efficiently optimized with minimal data and a small number of gradient steps. This allows for flexible adaptation to target model sizes while mitigating the limitations of manual initialization. Once $\mathcal{S}$ is trained, initialization is complete, and the model can be further trained \textbf{\textit{without}} additional constraints.

\renewcommand{\arraystretch}{0.83}
\begin{table*}
    \centering
    \setlength{\tabcolsep}{1.2 mm} 
        \caption{Performance of initializing models with variable depth on ImageNet-1K. ``Para.(M)'' denotes the number of parameters for each model size and ``FLOPs (G)'' represents the computational complexity. All models are trained 100K steps after initialization.}  
        \vspace{-0.1in}
        \resizebox{\textwidth}{!}{
        \begin{tabular}{@{}llccc|ccc|ccc|ccc|ccc@{}}
        \toprule[1.5pt]
        & & \multicolumn{3}{c}{DiT-B $L_{\text{4}}$} & \multicolumn{3}{c}{DiT-B $L_{\text{6}}$} & \multicolumn{3}{c}{DiT-B $L_{\text{8}}$} & \multicolumn{3}{c}{DiT-B $L_{\text{10}}$} & \multicolumn{3}{c}{DiT-B $L_{\text{12}}$}\\
        \cmidrule{3-17}
        & & \multicolumn{3}{c|}{\cellcolor{gray!15}{45.29 M / 14.56 G}} & \multicolumn{3}{c|}{\cellcolor{gray!15}{66.55 M / 21.82 G}} & \multicolumn{3}{c|}{\cellcolor{gray!15}{87.80 M / 29.09 G}} & \multicolumn{3}{c|}{\cellcolor{gray!15}{109.06 M / 36.36 G}} & \multicolumn{3}{c}{\cellcolor{gray!15}{130.32 M / 43.62 G}} \\
        \cmidrule{3-17}
        \multicolumn{2}{l}{Methods} & FID & sFID & IS & FID & sFID & IS & FID & sFID & IS & FID & sFID & IS & FID & sFID & IS \\
        % \midrule
        \toprule[1.1pt]
        \multirow{2}{*}{\rotatebox{90}{\small{\textbf{Direct}}}}
        & He-Init & 87.23 & 16.44 & 16.09 
                    & 80.37 & 16.49 & 17.20 
                    & 71.39 & 15.52 & 19.41
                    & 70.73 & 14.05 & 19.04
                    & 65.61 & 10.31 & 20.72 \\
        & Mimetic & 81.76 & 15.44 & 16.69 
                    & 79.87 & 17.55 & 18.58 
                    & 72.04 & 15.20 & 19.70
                    & 66.98 & 9.91 & 21.00
                    & 64.47 & 11.76 & 21.84 \\
        \midrule
        \multirow{4}{*}{\rotatebox{90}{\small{\textbf{Transfer}}}}
        & Share Init & 66.87 & 13.38 & 22.35 
                    & 59.03 & 14.00 & 24.61 
                    & 53.43 & 9.60 & 26.80
                    & 51.06 & 10.58 & 28.00
                    & 49.41 & 11.44 & 28.06 \\
        & LiGO & 62.04 & 14.14 & 23.55 
                    & 58.80 & 12.71 & 24.47 
                    & 53.22 & 13.43 & 27.54
                    & 54.25 & 12.88 & 27.28
                    & 52.60 & 10.49 & 27.34 \\
        & BK-SDM & 85.39 & 20.50 & 16.21 
                    & 63.44 & 13.34 & 22.99 
                    & 61.67 & 15.81 & 24.36
                    & 56.13 & 12.92 & 27.00
                    & 63.76 & 16.73 & 23.26 \\
        & Laptop-diff & 105.90 & 28.50 & 12.99 
                    & 68.71 & 20.06 & 21.00 
                    & 52.73 & 10.95 & 26.97
                    & 52.57 & 12.13 & 27.46
                    & 49.51 & 12.14 & 29.28 \\
        \midrule
        \multirow{4}{*}{\rotatebox{90}{\small{\textbf{Learngene}}}}
        & Heur-LG & 84.14 & 21.48 & 16.21 
                    & 70.84 & 16.84 & 19.94 
                    & 62.57 & 13.20 & 23.52
                    & 60.88 & 14.32 & 23.40
                    & 55.97 & 13.72 & 25.65 \\
        & Auto-LG & 81.63 & 19.81 & 18.18 
                    & 66.70 & 18.23 & 22.59 
                    & 64.07 & 12.87 & 24.09
                    & 59.80 & 11.39 & 25.32
                    & 56.94 & 10.76 & 26.06 \\
        & TLEG & 62.88 & 17.66 & 22.78 
                    & 54.97 & 14.32 & 26.76 
                    & 49.04 & 9.96 & 28.76
                    & 47.22 & 8.94 & 30.23
                    & 45.02 & 9.61 & 31.15 \\
        & \cellcolor{blue!12}{FINE} 
               & \cellcolor{blue!12}{\textbf{57.47}} & \cellcolor{blue!12}{\textbf{10.06}} & \cellcolor{blue!12}{\textbf{24.52}}
               & \cellcolor{blue!12}{\textbf{51.58}} & \cellcolor{blue!12}{\textbf{11.50}} & \cellcolor{blue!12}{\textbf{27.52}} 
               & \cellcolor{blue!12}{\textbf{45.34}} & \cellcolor{blue!12}{\textbf{7.18}} & \cellcolor{blue!12}{\textbf{30.46}} 
               & \cellcolor{blue!12}{\textbf{42.33}} & \cellcolor{blue!12}{\textbf{6.89}}
               & \cellcolor{blue!12}{\textbf{32.34}} 
               & \cellcolor{blue!12}{\textbf{42.74}} & \cellcolor{blue!12}{\textbf{8.45}} & \cellcolor{blue!12}{\textbf{31.55}} \\
             & & \textcolor{mygreen}{$\downarrow$4.57} 
                   & \textcolor{mygreen}{$\downarrow$3.32} 
                   & \textcolor{mygreen}{$\uparrow$0.97} 
                   & \textcolor{mygreen}{$\downarrow$3.39} 
                   & \textcolor{mygreen}{$\downarrow$1.21} 
                   & \textcolor{mygreen}{$\uparrow$0.76} 
                   & \textcolor{mygreen}{$\downarrow$3.70} 
                   & \textcolor{mygreen}{$\downarrow$2.42} 
                   & \textcolor{mygreen}{$\uparrow$1.70} 
                   & \textcolor{mygreen}{$\downarrow$4.89} 
                   & \textcolor{mygreen}{$\downarrow$2.05} 
                   & \textcolor{mygreen}{$\uparrow$2.11} 
                   & \textcolor{mygreen}{$\downarrow$2.28} 
                   & \textcolor{mygreen}{$\downarrow$1.16} 
                   & \textcolor{mygreen}{$\uparrow$0.40}\\
               
        % \midrule
        % \multirow{1}{*}{\rotatebox{90}{\small{\textbf{PT}}}}
        % &Direct PT & 300K\textcolor{red}{${\times n}$}  
        %         & 0 & 97.09 & 15.98 & 13.68 
        %             & 92.14 & 18.58 & 14.03
        %             & 87.10 & 21.16 & 14.90 
        %             & 81.36 & 24.24 & 16.12 
        %             & 74.67 & 13.33 & 17.09 \\
        \midrule[1.5pt]

        & & \multicolumn{3}{c}{DiT-L $L_{\text{4}}$} & \multicolumn{3}{c}{DiT-L $L_{\text{6}}$} & \multicolumn{3}{c}{DiT-L $L_{\text{8}}$} & \multicolumn{3}{c}{DiT-L $L_{\text{10}}$} & \multicolumn{3}{c}{DiT-L-$L_{\text{12}}$}\\
        \cmidrule{3-17}
        & & \multicolumn{3}{c|}{\cellcolor{gray!15}{80.05 M / 25.87 G}} & \multicolumn{3}{c|}{\cellcolor{gray!15}{117.83 M / 38.78 G}} & \multicolumn{3}{c|}{\cellcolor{gray!15}{155.61 M / 51.70 G}} & \multicolumn{3}{c|}{\cellcolor{gray!15}{193.38 M / 64.62 G}} & \multicolumn{3}{c}{\cellcolor{gray!15}{231.16 M / 77.53 G}} \\
        \cmidrule{3-17}
        \multicolumn{2}{l}{Methods} & FID & sFID & IS & FID & sFID & IS & FID & sFID & IS & FID & sFID & IS & FID & sFID & IS \\
        % \midrule
        \toprule[1.1pt]
        \multirow{2}{*}{\rotatebox{90}{\small{\textbf{Direct}}}}
        & He-Init & 78.46 & 20.81 & 17.98 
                    & 72.57 & 14.66 & 19.19 
                    & 64.91 & 15.16 & 20.84
                    & 59.64 & 10.85 & 22.61
                    & 58.99 & 9.36 & 23.47 \\
        & Mimetic & 77.55 & 14.95 & 18.15 
                    & 69.79 & 15.99 & 19.31 
                    & 64.45 & 11.55 & 22.13 
                    & 64.55 & 14.31 & 21.55 
                    & 62.09 & 13.29 & 22.70 \\

        \midrule
        \multirow{4}{*}{\rotatebox{90}{\small{\textbf{Transfer}}}}
        & Share Init & 58.54 & 12.38 & 24.89 
                    & 46.26 & 9.69 & 30.88 
                    & 43.76 & 8.37 & 32.25 
                    & 41.78 & 7.77 & 33.27 
                    & 40.62 & 9.38 & 34.45 \\

        & LiGO & 57.25 & 11.82 & 26.23 
                    & 52.87 & 10.56 & 27.89 
                    & 46.95 & 9.90 & 31.21 
                    & 47.45 & 11.12 & 31.54
                    & 45.96 & 9.19 & 31.76 \\
        & BK-SDM & 80.55 & 15.07 & 17.82 
                    & 58.09 & 13.10 & 26.29 
                    & 50.90 & 12.76 & 30.25 
                    & 51.47 & 10.91 & 30.24 
                    & 51.15 & 11.35 & 30.05 \\
        & Laptop-diff & 107.92 & 25.06 & 12.92 
                    & 63.02 & 11.90 & 23.33 
                    & 47.84 & 9.28 & 30.52 
                    & 47.50 & 11.60 & 31.09 
                    & 41.52 & 8.60 & 34.85 \\
        \midrule
        \multirow{4}{*}{\rotatebox{90}{\small{\textbf{Learngene}}}}
        & Heur-LG & 81.37 & 18.21 & 17.73 
                    & 65.49 & 14.64 & 22.96 
                    & 58.41 & 12.61 & 25.12 
                    & 55.34 & 12.43 & 26.58 
                    & 49.06 & 10.64 & 32.10 \\
        & Auto-LG & 77.66 & 15.95 & 18.73 
                    & 68.03 & 20.39 & 22.46 
                    & 60.42 & 15.22 & 25.92 
                    & 59.98 & 13.97 & 25.98 
                    & 53.37 & 10.72 & 27.39 \\
        & TLEG & 53.00 & 13.85 & 27.99 
                    & 46.69 & 8.92 & 30.85 
                    & 44.32 & 9.63 & 32.23 
                    & 41.15 & 9.98 & 34.80 
                    & 39.72 & 8.90 & 36.40 \\
        & \cellcolor{blue!12}{FINE}
               & \cellcolor{blue!12}{\textbf{48.72}} & \cellcolor{blue!12}{\textbf{8.77}} & \cellcolor{blue!12}{\textbf{28.27}}
               & \cellcolor{blue!12}{\textbf{44.38}} & \cellcolor{blue!12}{\textbf{8.86}} & \cellcolor{blue!12}{\textbf{31.41}} 
               & \cellcolor{blue!12}{\textbf{41.24}} & \cellcolor{blue!12}{\textbf{8.27}} & \cellcolor{blue!12}{\textbf{34.25}} 
               & \cellcolor{blue!12}{\textbf{36.53}} & \cellcolor{blue!12}{\textbf{6.92}}
               & \cellcolor{blue!12}{\textbf{36.79}} 
               & \cellcolor{blue!12}{\textbf{35.59}} & \cellcolor{blue!12}{\textbf{7.42}} & \cellcolor{blue!12}{\textbf{37.34}} \\
                & & \textcolor{mygreen}{$\downarrow$4.28} 
                   & \textcolor{mygreen}{$\downarrow$3.05} 
                   & \textcolor{mygreen}{$\uparrow$0.28} 
                   & \textcolor{mygreen}{$\downarrow$1.88} 
                   & \textcolor{mygreen}{$\downarrow$0.06} 
                   & \textcolor{mygreen}{$\uparrow$0.53} 
                   & \textcolor{mygreen}{$\downarrow$2.52} 
                   & \textcolor{mygreen}{$\downarrow$0.10} 
                   & \textcolor{mygreen}{$\uparrow$2.00} 
                   & \textcolor{mygreen}{$\downarrow$4.62} 
                   & \textcolor{mygreen}{$\downarrow$0.85} 
                   & \textcolor{mygreen}{$\uparrow$1.99} 
                   & \textcolor{mygreen}{$\downarrow$4.13} 
                   & \textcolor{mygreen}{$\downarrow$1.18} 
                   & \textcolor{mygreen}{$\uparrow$0.94}\\
        % \midrule
        % \multirow{1}{*}{\rotatebox{90}{\small{\textbf{PT}}}}
        % &Direct PT & 300K\textcolor{red}{${\times n}$} 
        %         & 0 & 92.25 & 24.28 & 13.88 
        %             & 83.71 & 14.54 & 15.54
        %             & 76.06 & 13.07 & 17.23 
        %             & 75.48 & 12.75 & 17.15 
        %             & 71.06 & 16.99 & 18.42 \\
        \bottomrule[1.5pt]
        \end{tabular}
        }
    \label{tab:lenth}
\vspace{-0.14in}
\end{table*}

\section{Experiments}
\label{sec:exper}
\paragraph{Datasets}
Our experiments focus on class-conditioned generation tasks, with FINE factorizing knowledge on ImageNet-1K~\cite{deng2009imagenet}.
To thoroughly assess the transferability, we further conduct experiments across diverse domains, including CelebA-HQ, LSUN-Bedroom, LSUN-Church, Hubble, MRI and Pokemon which differ significantly from the training dataset. Additional details are provided in Appendix~\ref{app:dataset}.

\vspace{-0.15in}
\paragraph{Basic Settings}
% \textbf{Details of Knowledge Factorization.}  
We adopt Diffusion Transformers (DiTs) as the backbone in our experiments, using two model variants: DiT-B and DiT-L. Both models operate with a latent patch size of $p=2$ and process $256 \times 256$ resolution images. 
For knowledge factorization, models are trained on ImageNet-1K for 300K steps with a batch size of 64 and a fixed learning rate of $1 \times 10^{-4}$, using AdamW on an NVIDIA RTX 4090 GPU. 
This factorization process is a \textbf{\textit{one-time cost}}, which extracts size-agnostic knowledge that can be efficiently transferred across models of varying sizes.

% \textbf{Details of Learngene Evaluation.} 
To evaluate the initialization capabilities of learngenes, we configure models with varying depths, ranging from $L_4$ to $L_{12}$. Each configuration is initialized and trained for 100K steps on ImageNet-1K under consistent conditions to assess initialization quality. Additionally, we perform experiments with DiT-B ($L_6$) and DiT-L ($L_6$) on downstream datasets to evaluate the transferability of learngenes.
The quality of generated images is assessed using Fr\'{e}chet Inception Distance~(FID)~\cite{heusel2017gans}, sFID~\cite{nash2021generating}, Fr\'{e}chet DINO distance~(FDD)~\cite{stein2023exposing} and Inception Score~\cite{salimans2016improved}. 
Further details are provided in Appendix~\ref{app:hyper}.

\vspace{-0.15in}
\paragraph{Baselines and State-of-the-art Methods}
Despite the success of diffusion models in image generation, research on their initialization remains limited. To fill this gap, we construct the first benchmark for evaluating learngene-based initialization in diffusion models and categorize state-of-the-art initialization approaches into three categories:
\textbf{(1)~Direct Initialization:} Methods like He-Init~\cite{chen2021empirical}, Mimetic Init~\cite{trockman2023mimetic} use heuristic rules or observed patterns to directly initialize networks. 
\textbf{(2)~Transfer Initialization:} Methods such as LiGO~\cite{wang2023learning} and Share Init~\cite{lan2019albert} directly adapt pre-trained parameters to target models, whereas Laptop-diff~\cite{zhang2024laptop} and BK-SDM~\cite{kim2024bk} perform cross-scale knowledge transfer via distillation and pruning.
\textbf{(3)~Learngene Initialization:} Methods such as Heur-LG~\cite{wang2022learngene}, Auto-LG~\cite{wang2023learngene}, and TLEG~\cite{xia2024transformer} extract and reuse compact, transferable neural network fragments for model initialization.
Further details of these methods are provided in Appendix~\ref{app:compare}.

\section{Results}
\subsection{Performance of FINE in Initializing Diffusion Models of Variable Sizes}
Practical deployment often requires diffusion models of different sizes to meet varying computational and memory constraints. 
To assess initialization performance under this setting, we construct 10 downstream models with varying depths and evaluate FINE against a range of state-of-the-art methods. 
As shown in Table~\ref{tab:lenth}, FINE consistently outperforms existing approaches across all model sizes, achieving notable FID reductions (up to 4.89 and 4.62) and IS gains (up to 2.11 and 1.99) for DiT-B and DiT-L (e.g., $L_{10}$), respectively, demonstrating strong scalability.

While direct initialization methods offer strong architectural compatibility, they consistently underperform compared to approaches leveraging pre-trained parameters, highlighting the value of partial knowledge transfer. 
However, excessive transfer can be detrimental—naively reusing full parameter sets often introduces structural misalignments and weakens adaptability. 
For instance, LiGO~\cite{wang2023learning}, which transfers all weights from a smaller model, tends to disrupt layer-wise coherence, resulting in suboptimal initialization for deeper architectures.

Distillation- and pruning-based approaches (e.g., Laptop-Diff~\cite{zhang2024laptop} and BK-SDM~\cite{kim2024bk}) provide structurally tolerant knowledge transfer but incur substantial overhead for each new model size, making them inefficient when scaling to multiple downstream models. In contrast, FINE achieves comparable adaptation with only a few hundred optimization steps. 
Furthermore, when the downstream model substantially differs in scale from the teacher (e.g., $L_4$), these methods degrade notably, as the compressed or distilled diffusion representations fail to preserve the layer-wise denoising hierarchy and temporal coherence essential to effective generative alignment, leading to suboptimal adaptation.

FINE achieves superior performance by disentangling reusable, size-agnostic knowledge from layer-specific variations through pre-training-based weight factorization, yielding consistently aligned shared representations and eliminating the heuristic layer selection required in prior learngene-based methods~\cite{wang2022learngene, xia2024transformer}. 
Its adaptive combination mechanism further enables flexible initialization across model scales, a critical advantage for image generation tasks that are highly sensitive to initialization quality.

\subsection{Performance Compared to Direct Pre-Training}
\begin{figure}[tb]
  \centering
  \includegraphics[width=\linewidth]{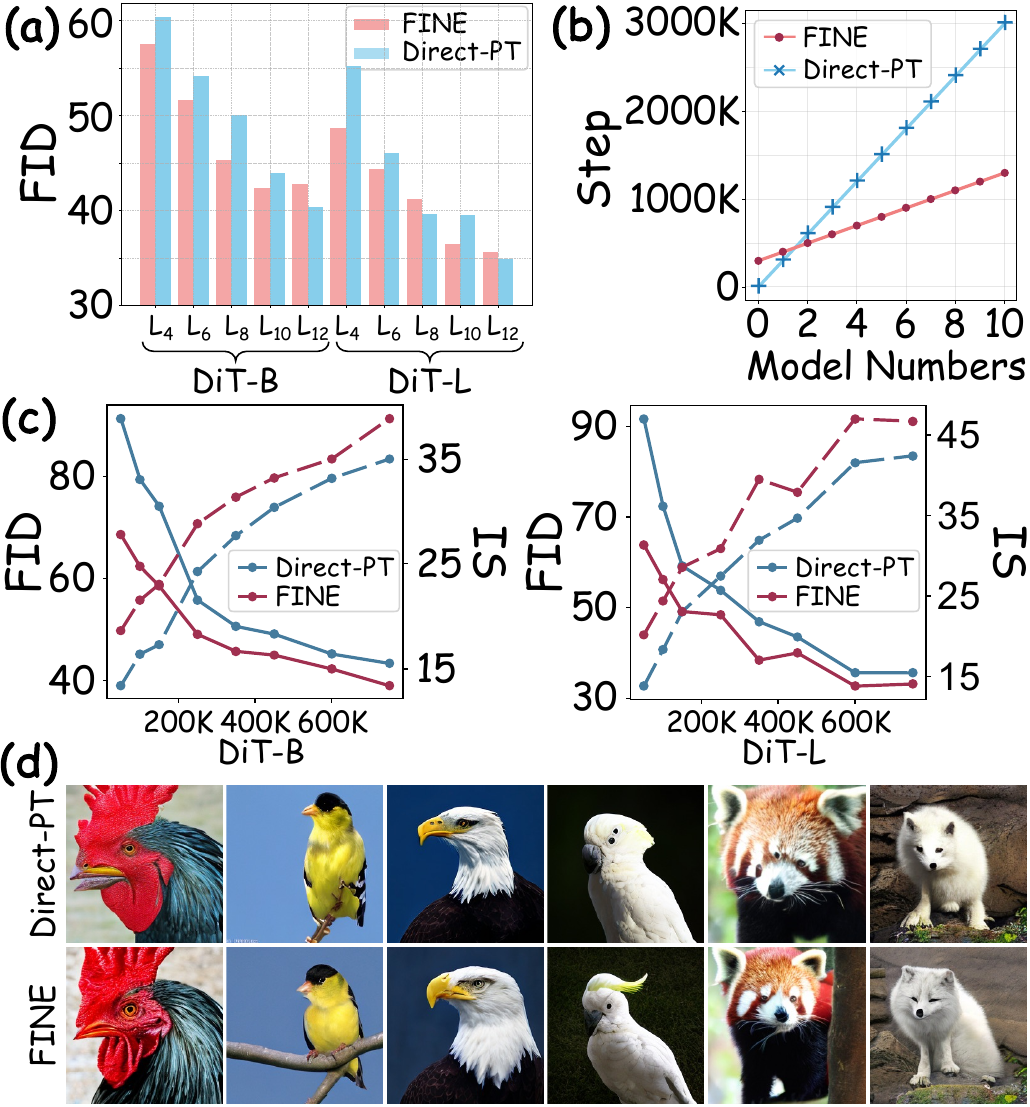}
  \vspace{-0.25in}
  \caption{Compared with Direct Pre-training. 
  (a)~Performance of models initialized by FINE and trained for \textbf{\textit{100K steps}} versus those directly pre-trained for \textbf{\textit{300K steps}} across 10 downstream model sizes. 
  (b)~Computational cost analysis as the number of initialized models increases.
  (c)~Training dynamics over 800K steps.
  % Detailed training process (800K steps) of models initialized by FINE and direct pre-training. 
  (d)~Visual comparison of samples generated by FINE-initialized and directly pre-trained models.
  }
  \label{fig:curve}
  \vspace{-0.15in}
\end{figure}

In scenarios lacking size-matched pre-trained models, FINE offers an efficient alternative by enabling direct initialization for variable-sized architectures. 
As shown in Figure~\ref{fig:curve}a, models initialized with FINE and trained for only 100K steps outperform those trained from scratch for 300K steps, demonstrating significant efficiency gains. This advantage scales with the number of target models: while direct pre-training requires $300K \times n$ steps for $n$ models, FINE reduces the total cost to $300K + 100K \times n$, achieving roughly a $3n\times$ speedup (Figure~\ref{fig:curve}b).

To analyze long-horizon training dynamics, we extend the budget to 800K steps. 
As shown in Figure~\ref{fig:curve}c, models initialized with FINE consistently achieve lower FID scores from the outset, benefiting from size-agnostic knowledge encapsulated in shared learngenes. Notably, FINE reaches comparable FID up to 200K steps earlier than models trained from scratch, reflecting enhanced convergence stability.
These results demonstrate that the one-time cost of pre-training for knowledge factorizatable models (Section~\ref{sec:KF}) yields lasting downstream benefits, with the persistent performance gap confirming that initialization quality influences the entire training trajectory—supporting early-stage metrics (Table~\ref{tab:lenth}) as reliable indicators of initialization effectiveness.

Figure~\ref{fig:curve}d further visualizes generation quality, showing that FINE-initialized models achieve superior semantic coherence and structural consistency, underscoring FINE’s effectiveness in stabilizing generative alignment.

\subsection{Performance of FINE in Initializing Models on Downstream Datasets}
\renewcommand{\arraystretch}{0.88}
\begin{table*}[t]
    \centering
    \setlength{\tabcolsep}{1.6 mm}
    \caption{Performance of model initialization on diverse downstream datasets. 
    FID is used for natural image datasets (first three), while FDD is used for non-natural ones (last three).}
    \vspace{-0.13in}
    \resizebox{\textwidth}{!}{
        \begin{tabular}{@{}llcc|cc|cc|cc|cc|cc@{}}
        \toprule[1.5pt]
             & & \multicolumn{2}{c}{CelebA} & \multicolumn{2}{c}{Bedroom} & \multicolumn{2}{c}{Church} & \multicolumn{2}{c}{Hubble} & \multicolumn{2}{c}{MRI} & \multicolumn{2}{c}{Pokemon}\\
             \cmidrule(r){3-8}
             \cmidrule(l){9-14}
             \multicolumn{2}{l}{Methods} & DiT-B & DiT-L & DiT-B & DiT-L & DiT-B & DiT-L & DiT-B & DiT-L & DiT-B & DiT-L & DiT-B & DiT-L \\
             \midrule[1.1pt]
             \multirow{2}{*}{\rotatebox{90}{\small{\textbf{Direct}}}}
             & He-Init 
             & 18.57 & 14.55 & 42.90 & 32.88 & 41.01 & 24.27
             & 0.320 & 0.235 & 0.170 & 0.119 & 0.897 & 0.925\\
             & Mimetic 
             & 16.87 & 11.66 & 30.48 & 29.66 & 33.65 & 25.00 
             & 0.281 & 0.271 & 0.180 & 0.111 & 0.902 & 0.920\\
             \cmidrule(r){1-8}
             \cmidrule(l){9-14}
             \multirow{4}{*}{\rotatebox{90}{\small{\textbf{Transfer}}}}
             & Share Init 
             & 9.11 & 9.40 & 25.47 & 17.90 & 22.49 & 19.88 
             & 0.190 & 0.119 & 0.057 & 0.047 & 0.463 & 0.421\\
             & LiGO 
             & 11.90 & 15.92 & 28.87 & 26.01 & 37.56 & 33.63
             & 0.180 & 0.164 & 0.064 & 0.079 & 0.515 & 0.558\\
             & BK-SDM 
             & 10.37 & 15.71 & 34.02 & 18.39 & 27.30 & 21.10 
             & 0.229 & 0.140 & 0.058 & 0.054 & 0.454 & 0.482\\
             & Laptop-diff
             & 12.62 & 10.52 & 22.85 & 27.01 & 24.73 & 25.19 
             & 0.153 & 0.141 & 0.063 & 0.051 & 0.466 & 0.478\\
             \cmidrule(r){1-8}
             \cmidrule(l){9-14}
             \multirow{4}{*}{\rotatebox{90}{\small{\textbf{Learngene}}}}
             & Heur-LG 
             & 13.23 & 10.84 & 36.98 & 24.42 & 29.13 & 17.09 
             & 0.293 & 0.314 & 0.127 & 0.099 & 0.865 & 0.919\\
             & Auto-LG 
             & 15.02 & 16.54 & 46.56 & 38.98 & 44.15 & 31.58 
             & 0.302 & 0.270 & 0.110 & 0.148 & 0.705 & 0.764\\
             & TLEG 
             & 8.27 & 10.91 & 20.43 & 19.43 & 19.30 & 18.29
             & 0.226 & 0.124 & 0.057 & 0.052 & 0.428 & 0.412\\
             & \cellcolor{blue!12}{FINE} 
             & \cellcolor{blue!12}{\textbf{7.99}} & \cellcolor{blue!12}{\textbf{8.41}} & \cellcolor{blue!12}{\textbf{17.83}} & \cellcolor{blue!12}{\textbf{14.90}} & \cellcolor{blue!12}{\textbf{17.29}} & \cellcolor{blue!12}{\textbf{15.80}} & \cellcolor{blue!12}{\textbf{0.119}} & 
             \cellcolor{blue!12}{\textbf{0.101}} & 
             \cellcolor{blue!12}{\textbf{0.049}} & 
             \cellcolor{blue!12}{\textbf{0.041}} & 
             \cellcolor{blue!12}{\textbf{0.407}} &
             \cellcolor{blue!12}{\textbf{0.380}}\\
             & & \textcolor{mygreen}{$\downarrow$0.28} 
               & \textcolor{mygreen}{$\downarrow$0.24} 
               & \textcolor{mygreen}{$\downarrow$2.60} 
               & \textcolor{mygreen}{$\downarrow$3.00} 
               & \textcolor{mygreen}{$\downarrow$2.01} 
               & \textcolor{mygreen}{$\downarrow$1.29}
            & \textcolor{mygreen}{$\downarrow$0.029} 
               & \textcolor{mygreen}{$\downarrow$0.018} 
               & \textcolor{mygreen}{$\downarrow$0.008} 
               & \textcolor{mygreen}{$\downarrow$0.006} 
               & \textcolor{mygreen}{$\downarrow$0.011} 
               & \textcolor{mygreen}{$\downarrow$0.032}\\
           \cmidrule(r){1-8}
           \cmidrule(l){9-14}
           \multirow{1}{*}{\rotatebox{90}{\small{\textbf{PT}}}} 
           & Full FT 
           & 9.97 & 8.65 & 24.43 & 19.58 & 20.65 & 19.10
           & 0.148 & 0.124 & 0.060 & 0.048 & 0.418 & 0.421\\
           \bottomrule[1.5pt]
        \end{tabular}
        }
    \label{tab:downstream}
    \vspace{-0.15in}
\end{table*}

The learngenes extracted through knowledge factorization are not only size-agnostic, but also domain-agnostic to a certain extent. As shown in Table~\ref{tab:downstream}, FINE flexible model initialization across diverse downstream datasets and consistently outperforms other initialization methods.

As observed previously, transferring parameters from pre-trained models offers a more direct and effective strategy than rule-based or pattern-based initialization.
Notably, models initialized by FINE outperform those fine-tuned directly from pre-trained models, while transferring only 35\% of the parameters. 
This reinforces the notion that transferring more parameters does not always result in better performance~\cite{feng2024transferring}, particularly when significant gaps exist between downstream tasks (e.g., Hubble and MRI) and training tasks, as excessive redundant knowledge can hinder the model's adaptability.

\subsection{Accelerated Convergence Achieved by FINE}
\begin{figure}[tb]
  \centering
  \includegraphics[width=\linewidth]{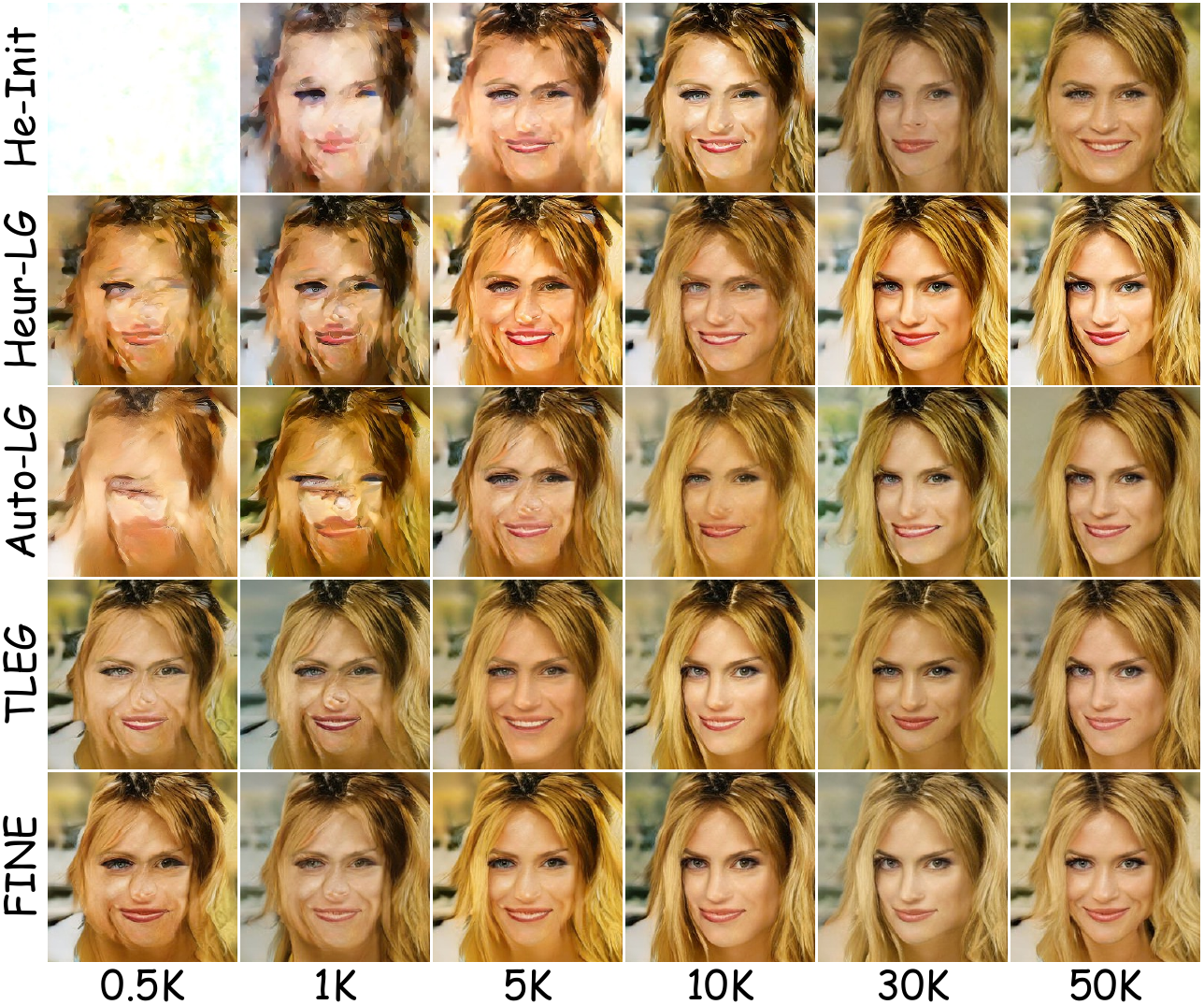}
  \vspace{-0.3in}
  \caption{Visualization of convergence speed of FINE and other learngene-based methods on CelebA-HQ.}
  \vspace{-0.2in}
  \label{fig:converg}
\end{figure}
Initialization plays a crucial role in determining model convergence speed~\cite{zhu2021gradinit, narkhede2022review}. Prior studies~\cite{wang2022learngene, xia2024transformer} show that transferring shared knowledge through learngenes can significantly accelerate downstream adaptation compared to training from scratch.

Figure~\ref{fig:converg} presents a visual comparison of convergence progress on CelebA-HQ~\cite{huang2018introvae}, with images generated at different training steps. 
While all methods benefit from knowledge transfer, FINE achieves faster convergence and generates higher-quality outputs at earlier stages. 
This demonstrates its ability to provide effective, task-adaptive initialization, further reinforcing the general utility of learngenes across diverse domains.

% \subsection{Extension to Classification Tasks and Other Architectures}
\subsection{Extension to Classification Tasks}
\begin{table*}[t]
    \centering
    \setlength{\tabcolsep}{2.0 mm} 
    \caption{Performance on the classification task (e.g., ImageNet-1K) using DeiT-Ti and DeiT-S of varying depths. ``Para.(M)'' denotes the \textit{average} number of parameters transferred during initialization.}
    \vspace{-0.1in}
    \resizebox{\linewidth}{!}{
        \begin{tabular}{@{}llcccccc|cccccc@{}}
        \toprule[1.5pt]
        & & & \multicolumn{5}{c}{DeiT-Ti} & & \multicolumn{5}{c}{DeiT-S}\\
        \cmidrule{4-8}
        \cmidrule{10-14}
        \multicolumn{2}{l}{Methods} & Para. & $L_4$ & $L_6$ & $L_8$ & $L_{10}$ & $L_{12}$ & Para. & $L_4$ & $L_6$ & $L_8$ & $L_{10}$ & $L_{12}$ \\
        \midrule[1.1pt]
        \multirow{3}{*}{\rotatebox{90}{\small{\textbf{Direct}}}}
        & He-Init~\cite{chen2021empirical} 
                & \cellcolor{gray!15}{0} & 34.73 & 40.60 & 43.67 & 46.84 & 48.28 
                & \cellcolor{gray!15}{0} & 42.20 & 49.35 & 52.14 & 53.68 & 55.51 \\
        & Mimetic~\cite{trockman2023mimetic}
                & \cellcolor{gray!15}{0} & 35.07 & 40.18 & 43.18 & 46.29 & 48.05 
                & \cellcolor{gray!15}{0} & 43.29 & 49.06 & 53.00 & 54.13 & 55.58\\
        & GHN-3~\cite{knyazev2023canwescale}
                & \cellcolor{gray!15}{0} & 40.92 & 44.97 & 46.56 & 49.05 & 48.87 
                & \cellcolor{gray!15}{0} & 45.37 & 48.98 & 50.15 & 52.35 & 53.19\\
        \midrule
        \multirow{2}{*}{\rotatebox{90}{\small{\textbf{Trans.}}}}
        & Share Init~\cite{lan2019albert}
                   & \cellcolor{gray!15}{0.8} & 55.16 & 59.83 & 62.52 & 64.25 & 65.33 
                   & \cellcolor{gray!15}{2.5} & 64.95 & 69.66 & 71.65 & 72.65 & 73.34\\
        & LiGO~\cite{wang2023learning}
             & \cellcolor{gray!15}{2.2} & --- & 58.98 & 60.18 & 59.85 & 60.91 
             & \cellcolor{gray!15}{7.9} & --- & 68.57 & 69.88 & 69.74 & 69.97\\
        \midrule
        \multirow{4}{*}{\rotatebox{90}{\small{\textbf{Learngene}}}}
        & Heur-LG~\cite{wang2022learngene}
                & \cellcolor{gray!15}{1.7} & 41.47 & 47.37 & 50.51 & 53.55 & 55.52 
                & \cellcolor{gray!15}{6.1} & 52.33 & 57.32 & 61.67 & 64.35 & 65.89\\
        & Auto-LG~\cite{wang2023learngene}
                & \cellcolor{gray!15}{2.2} & 52.38 & 61.80 & 64.56 & 65.88 & 66.79 
                & \cellcolor{gray!15}{7.9} & 63.19 & 70.50 & 72.19 & 73.29 & 73.81\\
        & TLEG~\cite{xia2024transformer}
             & \cellcolor{gray!15}{1.3} & 55.00 & 60.50 & 62.88 & 64.40 & 65.40 
             & \cellcolor{gray!15}{4.3} & 65.43 & 70.52 & 72.14 & 73.15 & 73.84\\
        & \cellcolor{blue!12}{FINE} 
               & \cellcolor{blue!12}{1.4} & \cellcolor{blue!12}{\textbf{57.88}} & \cellcolor{blue!12}{\textbf{62.52}} & \cellcolor{blue!12}{\textbf{64.83}} & \cellcolor{blue!12}{\textbf{66.23}} & \cellcolor{blue!12}{\textbf{67.02}}
               & \cellcolor{blue!12}{4.4} & \cellcolor{blue!12}{\textbf{68.47}} & \cellcolor{blue!12}{\textbf{72.47}} & \cellcolor{blue!12}{\textbf{73.76}} & \cellcolor{blue!12}{\textbf{74.49}} & \cellcolor{blue!12}{\textbf{74.75}}\\
               & & & \textcolor{mygreen}{$\uparrow$2.72} 
                   & \textcolor{mygreen}{$\uparrow$0.72} 
                   & \textcolor{mygreen}{$\uparrow$0.27} 
                   & \textcolor{mygreen}{$\uparrow$0.35} 
                   & \textcolor{mygreen}{$\uparrow$0.23} 
               & & \textcolor{mygreen}{$\uparrow$3.04} 
                   & \textcolor{mygreen}{$\uparrow$1.95} 
                   & \textcolor{mygreen}{$\uparrow$1.57} 
                   & \textcolor{mygreen}{$\uparrow$1.20} 
                   & \textcolor{mygreen}{$\uparrow$0.91}\\
        \bottomrule[1.5pt]
        \end{tabular}   
        }
    \label{tab:lenth_DeiT}
    \vspace{-0.2in}
\end{table*}
To assess the generality of FINE beyond diffusion models, we apply it to classification tasks using DeiT~\cite{touvron2021training}. As shown in Table~\ref{tab:lenth_DeiT}, FINE achieves consistently strong performance across models with different depths, without requiring architectural modifications or additional heuristics.

Compared to methods like LiGO, which introduce random transformations that may compromise stability, FINE employs a deterministic recomposition of shared components and lightweight tuning of $\Sigma_{\star}^{(l)}$ for efficient adaptation.
Moreover, FINE extends the utility of learngenes beyond layer-specific reuse by introducing cross-layer factorization. This design enables task-agnostic, architecture-robust knowledge transfer, underscoring the flexibility and reusability of learngenes across diverse learning scenarios.

\subsection{Ablation and Analysis}
% \subsubsection{Existence of Size-agnostic Knowledge}
% \label{sec:exist}
% \begin{figure}[tb]
%   \centering
%   \includegraphics[width=0.9\linewidth]{image/FINE_agnos.pdf}
%   \vspace{-0.1in}
%   \caption{t-SNE of singular vectors in layers $L_1$ to $L_6$ of the pre-trained DiT-B. Each layer is represented by a distinct color, while the yellow stars indicate the clustering centers. }
%   \label{fig:exist}
%   % \vspace{-0.2in}
% \end{figure}
% FINE identifies the shared singular vectors across layers in DiT, derived via SVD, as size-agnostic knowledge, enabling the efficient initialization of models with varying sizes, especially those with different depths. To visually confirm the presence of this size-agnostic knowledge in DiT, we perform SVD on the pre-trained model and cluster the singular vectors across layers, as illustrated in Figure~\ref{fig:exist}.

% The results demonstrate significant similarity among certain singular vectors across layers, forming clusters of varying sizes. These clustering centers indicate the existence of the size-agnostic knowledge shared across layers. Consequently, learngenes are trained to capture this size-agnostic knowledge by updating under the constraint in Eq.~\eqref{equ:svd_sip}, effectively encapsulating size-agnostic knowledge while isolating size-specific information.

\subsubsection{Ablation on Knowledge Factorization}  
To demonstrate the advantages of size-agnostic knowledge, we ablate knowledge factorization by independently applying SVD to each layer’s weight matrix, selecting the top singular vectors in $U$ and $V$ to match the number of extracted learngenes.
The ablation results are presented in Table~\ref{tab:ab1}.

\renewcommand{\arraystretch}{0.85}
\begin{table}[t]
    \centering
    \setlength{\tabcolsep}{1.6 mm}
    \caption{Ablation study on knowledge factorization.}
    \vspace{-0.1in}
    \resizebox{\linewidth}{!}{
        \begin{tabular}{@{}lccc|ccc@{}}
        \toprule[1.5pt]
             & \multicolumn{3}{c}{DiT-B $L_{6}$} & \multicolumn{3}{c}{DiT-L $L_{6}$}\\
             \cmidrule(r){2-4}
             \cmidrule(l){5-7}
             & FID & sFID & IS & FID & sFID & IS \\
             \midrule[1.1pt]
             From Scratch & 80.37 & 16.49 & 17.20 & 72.57 & 14.66 & 19.19 \\
             w/o Factorize & 62.86 & 12.86 & 23.88 & 56.42 & 13.96 & 26.84 \\
             \midrule
             \cellcolor{blue!12}{FINE} & \cellcolor{blue!12}{\textbf{51.58}} & \cellcolor{blue!12}{\textbf{11.50}} & \cellcolor{blue!12}{\textbf{27.52}} & \cellcolor{blue!12}{\textbf{44.38}} & \cellcolor{blue!12}{\textbf{8.86}} & \cellcolor{blue!12}{\textbf{31.41}} \\
             & \textcolor{mygreen}{$\downarrow$11.28} & \textcolor{mygreen}{$\downarrow$1.36} & \textcolor{mygreen}{$\uparrow$3.64} & \textcolor{mygreen}{$\downarrow$12.04} & \textcolor{mygreen}{$\downarrow$5.10} & \textcolor{mygreen}{$\uparrow$4.57}\\
             \bottomrule[1.5pt]
        \end{tabular}
        }
    \label{tab:ab1}
    \vspace{-0.05in}
\end{table}

The results demonstrate that applying SVD independently to each weight matrix enables limited knowledge transfer, but yields highly layer-specific components with poor reusability.
In contrast, FINE factorizes knowledge into shared, size-agnostic components that can be flexibly recombined for initializing models of different sizes.
While the factorization incurs a higher initial cost, it is a \textbf{\textit{one-time cost}} that becomes negligible when reused across multiple models (Figure~\ref{fig:curve}b), making FINE a scalable and efficient solution for initialization and transfer.

\subsubsection{Effect of Adaptable Manual-Free Initialization}
\begin{table}[t]
    \centering
    \setlength{\tabcolsep}{1.3 mm}
    \caption{Ablation study on the initialization of $\Sigma$.}
    \vspace{-0.1in}
    \resizebox{\linewidth}{!}{
        \begin{tabular}{@{}lccc|ccc@{}}
        \toprule[1.5pt]
             & \multicolumn{3}{c}{DiT-B $L_{12}$} & \multicolumn{3}{c}{DiT-L $L_{12}$}\\
             \cmidrule(r){2-4}
             \cmidrule(l){5-7}
             & FID & sFID & IS & FID & sFID & IS \\
             \midrule[1.1pt]
             Random Init. & 77.70 & 17.01 & 18.46 & 73.58 & 15.97 & 18.80 \\
             Identical Init. & 47.84 & 11.24 & 28.55 & 42.53 & 10.27 & 32.04  \\
             Linear Init. & 46.71 & 9.22 & 29.49 & 39.34 & 7.53 & 32.52  \\
             \midrule
             \cellcolor{blue!12}{Trainable Init.} & \cellcolor{blue!12}{\textbf{42.74}} & \cellcolor{blue!12}{\textbf{8.45}} & \cellcolor{blue!12}{\textbf{31.55}} & \cellcolor{blue!12}{\textbf{35.59}} & \cellcolor{blue!12}{\textbf{7.42}} & \cellcolor{blue!12}{\textbf{37.34}} \\
             & \textcolor{mygreen}{$\downarrow$3.97} & \textcolor{mygreen}{$\downarrow$0.77} & \textcolor{mygreen}{$\uparrow$2.06} & \textcolor{mygreen}{$\downarrow$3.75} & \textcolor{mygreen}{$\downarrow$0.11} & \textcolor{mygreen}{$\uparrow$4.82}\\
             \bottomrule[1.5pt]
        \end{tabular}
        }
    \label{tab:KF}
    \vspace{-0.15in}
\end{table}

We evaluate the benefits of FINE's manual-free model initialization by comparing its adaptable approach with conventional rule-based methods.  
Table~\ref{tab:KF} presents the performance of models initialized with learngenes using different strategies for initializing $\Sigma$.

Results indicate that the size-agnostic knowledge encapsulated in learngenes $\mathcal{G}$ can be recombined across model sizes via rule-based schemes such as identical~\cite{lan2019albert} or linear initialization~\cite{xia2024transformer}. 
Although rule-based initialization outperforms random initialization, they lack flexibility for model-specific adaptation.
FINE addresses this by introducing a lightweight, trainable $\Sigma$ within a compact parameter space. With minimal data, $\Sigma$ can be optimized to customize initialization for each model size, fully leveraging the generality of $\mathcal{G}$ for effective transfer.

\subsubsection{Visualization of $\Sigma$ Across Models of Variable Sizes}
\begin{figure}[tb]
  \centering
  \includegraphics[width=\linewidth]{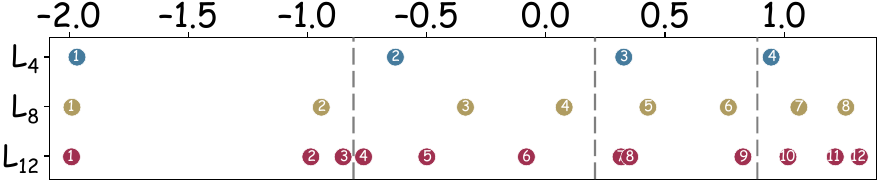}
  \vspace{-0.25in}
  \caption{Visualization of the relationship between layer positions (e.g., DiT-L) and corresponding values of $\Sigma$ after applying PCA. The white number $i$ on each dot indicates the $i$-th layer.}
  \label{fig:sigma}
  \vspace{-0.1in}
\end{figure}
We further visualize the values of $\Sigma$ across layers in models of varying sizes during learngene initialization. Specifically, for each layer, we concatenate $\Sigma_{\star}^{(l)}\in \mathcal{S}$ and apply PCA~\cite{abdi2010principal} to enhance interpretability, as shown in Figure~\ref{fig:sigma}.

The visualization reveals a clear linear relationship in $\Sigma_{\star}^{(l)}$ across model layers, consistent with observations in~\cite{xia2024transformer}, as evidenced by the nearly equal spacing between points of the same color. 
Notably, layers from smaller models align with corresponding segments in deeper models (e.g., the first layer of DiT-L $L_4$ aligns with the first two layers of $L_8$, and so on), suggesting strong cross-scale structural coherence.

\section{Conclusion}
We introduce FINE, a novel initialization method for variable-sized diffusion models, particularly suited for scenarios without size-matched pre-trained models. By factorizing weight matrices into size-agnostic components shared across layers, FINE enables flexible and efficient initialization through lightweight adaptation.
FINE is the first learngene framework applied to diffusion models and image generation, and the first to accelerate diffusion training via structured decomposition. Beyond diffusion, FINE generalizes effectively to classification tasks and alternative architectures, demonstrating the robustness and transferability of learngenes. Extensive experiments validate its superior performance across diverse tasks and model scales.

\section*{Acknowledgement}
We sincerely appreciate Freepik for contributing to the figure design. 
This research was supported by the Jiangsu Science Foundation (BG2024036, BK20243012), the National Natural Science Foundation of China (625B2045, 62125602, U24A20324, 92464301, 62306073), the New Cornerstone Science Foundation through the XPLORER PRIZE, the Fundamental Research Funds for the Central Universities (2242025K30024), and SEU Innovation Capability Enhancement Plan for
Doctoral Students (CXJH\_SEU 26023).

{
    \small
    \bibliographystyle{ieeenat_fullname}
    \bibliography{main}
}


\clearpage
\appendix
\setcounter{page}{1}
\maketitlesupplementary

\section{Training Details}
\label{app:detail}
\subsection{Details of Knowledge Factorization}
\label{app:condense}
Algorithm~\ref{alg:algorithm} outlines the pseudo code for factorizing knowledge and encapsulating size-agnostic knowledge into shared singular vectors $U_{\star}$ and $V_{\star}$, referred to as learngenes.

\begin{algorithm}[h]
    \caption{Knowledge Factorization for Encapsulating Size-agnostic Knowledge}
    \small
    \label{alg:algorithm}
    \textbf{Input}: A DiT model $\varepsilon_{\theta}$ with $L$ layers, training dataset $\{(z^{(i)}, c^{(i)})\}_{i=1}^m$, number of epochs $N_{\text{ep}}$, batch size $B$, learning rate $\alpha$\\
    \textbf{Output}: Shared singular vectors $\mathcal{U}$ and $\mathcal{V}$ (i.e., learngene) encapsulating size-agnostic knowledge
    \begin{algorithmic}[1]
        \STATE Random initialize the set of shared singular vectors $\mathcal{U}$ and $\mathcal{V}$, and layer-specific singular values $\mathcal{S}$
        \STATE Construct weight matrices $\theta$ using $\mathcal{U}$, $\mathcal{V}$ and $\mathcal{S}$ according to Eq.~\eqref{equ:svd_sip}
        \FOR{$ep = 1$ to $N_{\text{ep}}$}
            \FOR{each batch $\{(z^{(i)}, c^{(i)})\}_{i=1}^B$}
                \FOR{each $U_{\star}$, $V_{\star}$ and $\Sigma_{\star}^{(l)}$ in $\mathcal{U}$, $\mathcal{V}$ and $\mathcal{S}$}
                    \STATE Update $W_{\star}^{(l)}$ in $\theta$ based on the current values of $U_{\star}$, $V_{\star}$ and $\Sigma_{\star}^{(l)}$ under the rule of Eq.~\eqref{equ:svd}
                \ENDFOR
                \STATE Perform a forward propagate $\varepsilon_{\theta}(z_t, t, c)$ through the model for each noisy input $z_t$ at timestep $t$
                \STATE Calculate $\mathcal{L}_{\text{batch}}=\frac{1}{B} \sum_{i=1}^B \mathcal{L}(\varepsilon, \varepsilon_{\theta}(z_t^{(i)}, c^{(i)}, t))$ according to Eq.~\eqref{eq:loss}
                \STATE Backward propagate the loss $\mathcal{L}_{\text{batch}}$ to compute the gradients with respect to $\mathcal{U}$, $\mathcal{V}$ and $\mathcal{S}$: $\nabla_{\mathcal{U}} \mathcal{L}_{\text{batch}}, \nabla_{\mathcal{V}} \mathcal{L}_{\text{batch}}$ and $\nabla_{\mathcal{S}} \mathcal{L}_{\text{batch}}$
                \STATE Update $\mathcal{U}$, $\mathcal{V}$ and $\mathcal{S}$\\ \quad $\mathcal{U} := \mathcal{U} - \alpha \cdot \nabla_{\mathcal{U}} \mathcal{L}_{\text{batch}}$\\ \quad
                $\mathcal{V} := \mathcal{V} - \alpha \cdot \nabla_{\mathcal{V}} \mathcal{L}_{\text{batch}}$\\ \quad
                $\mathcal{S} := \mathcal{S} - \alpha \cdot \nabla_{\mathcal{S}} \mathcal{L}_{\text{batch}}$
            \ENDFOR
        \ENDFOR
    \end{algorithmic}
\end{algorithm}

\subsection{Details of Downstream Datasets}
\label{app:dataset}
Table~\ref{tab:datasets} provides an overview of six downstream datasets: CelebA-HQ~\cite{huang2018introvae}, LSUN-Bedroom, LSUN-Church~\cite{wang2017knowledge}, Hubble, MRI and Pokemon.
LSUN-Bedroom and LSUN-Church are subsets of the Large-Scale Scene Understanding (LSUN) dataset~\cite{wang2017knowledge}, containing scene images of bedrooms and churches, respectively, with a resolution of $256 \times 256$ pixels.
CelebA-HQ is a high-quality variant of the CelebA dataset~\cite{liu2018large}, featuring large-scale facial images of celebrities, resized to $256 \times 256$ pixels.
\begin{table}[h]
    \centering
    \setlength{\tabcolsep}{7 mm}
    % \vspace{-0.15in}
    \caption{Characteristics of downstream datasets.}
    % \vspace{0.05in}
    \resizebox{\linewidth}{!}{
        \begin{tabular}{@{}lcc@{}}
        \toprule[1.1pt]
        \textbf{Dataset} & \textbf{Total} & \textbf{Resolution} \\
        \cmidrule[1.1pt]{1-3}
        CelebA & 30,000 & 256$\times$256  \\
        LSUN-Bedroom & 3,033,042 & 256$\times$256 \\
        LSUN-Church & 126,227 & 256$\times$256 \\
        Hubble & 2706 & 256$\times$256  \\
        MRI & 3753 & 256$\times$256 \\
        Pokemon & 833 & 256$\times$256 \\
        \bottomrule[1.1pt]
        \end{tabular}
        }
    \label{tab:datasets}
    % \vspace{-0.05in}
\end{table}

\subsection{Hyper-parameters}
\label{app:hyper}
Table~\ref{tab:hyper_main} and Table~\ref{tab:hyper_down} present the basic settings, including batch size, training steps, optimizer and other settings for FINE encapsulating size-agnostic knowledge into shared singular vectors $U_{\star}$ and $V_{\star}$ and training the models initialized with learngenes on various datasets, respectively.

\begin{table}
    \centering
    \setlength{\tabcolsep}{4.6 mm}
    % \vspace{-0.15in}
    \caption{Hyper-parameters for FINE factorizing knowledge on ImageNet-1K.}
    \resizebox{\linewidth}{!}{
    % \vspace{0.05in}
        \begin{tabular}{@{}lr@{}}
        \toprule[1.1pt]
        \textbf{Training Settings} & \textbf{Configuration} \\
        \midrule[1pt]
        optimizer & AdamW\\
        learning rate & 1e-4\\
        weight decay & 0\\
        optimizer momentum & 0.9\\
        batch size & 64\\
        training steps & 300K\\
        drop path & 0.5\\
        sigma share & B: 100 $\mid$ L:148\\
        class dropout & 0.1\\
        vae & stabilityai / sd-vae-ft-ema \\
        \bottomrule[1.1pt]
        \end{tabular}
        }
    \label{tab:hyper_main}
    % \vspace{-0.15in}
\end{table}

\begin{table*}[htb]
    \centering
    \setlength{\tabcolsep}{4 mm}
    \caption{Hyper-parameters for neural networks trained on downstream datasets.}
    \resizebox{\textwidth}{!}{
        \begin{tabular}{@{}lcccccc@{}}
        \toprule[1.1pt]
        \textbf{Dataset} & \textbf{Batch Size} & \textbf{Training Steps} & \textbf{Learning Rate} & \textbf{Drop Last} & \textbf{Droppath Rate} & \textbf{Optimizer}\\
        \midrule[1pt]
        CelebA & 64 & 100K & 1e-4 & True & 0.1 & AdamW\\
        Bedroom & 64 & 150K & 1e-4 & True & 0.1  & AdamW\\
        Church & 64 & 150K & 1e-4 & True & 0.1 & AdamW\\
        Hubble & 64 & 20K & 1e-4 & True & 0.1 & AdamW\\
        MRI & 64 & 20K & 1e-4 & True & 0.1  & AdamW\\
        Pokemon & 64 & 20K & 1e-4 & True & 0.1 & AdamW\\
        \bottomrule[1.1pt]
        \end{tabular}
        }
    \label{tab:hyper_down}
\end{table*}

\subsection{Compared methods}
\label{app:compare}
\paragraph{Direct Initialization.}
Models are initialized using predefined rules (e.g., He-Init~\cite{chen2021empirical}) or observed patterns (e.g., Mimetic Init~\cite{trockman2023mimetic}).

\vspace{-0.1in}
\paragraph{Conventional Knowledge Transfer.}
These methods focus on transferring knowledge from pre-trained models to new ones. 
For example, LiGO~\cite{wang2023learning} trains larger model by leveraging knowledge from a pre-trained smaller one,
while Share init~\cite{lan2019albert} reuses trained blocks across multiple layers to initialize models with variable depths.
Laptop-Diff~\cite{zhang2024laptop} and BK-SDM~\cite{kim2024bk} adopt a prune-then-distill strategy, in which the pre-trained model is first layer-wise pruned to match the target architecture, followed by knowledge distillation to recover performance.

\vspace{-0.1in}
\paragraph{Learngene-Based Methods.}
We adapt existing learngene methods to diffusion models in this paper. 
Heur-LG~\cite{wang2022learngene} selects layers with minimal gradient changes during training as the learngenes, while Auto-LG~\cite{wang2023learngene} employs meta-learning to identify layers in the pre-trained model that share similar representations to those required by downstream tasks.
TLEG~\cite{xia2024transformer} builds on the linear relationships observed between different layers of transformer architectures.

These methods provide diverse strategies for initializing diffusion models, each with varying reliance on prior knowledge and pre-learned patterns, advancing the state-of-the-art in model initialization.

\section{Additional Results}
\begin{figure}[t]
  \centering
  \includegraphics[width=0.9\linewidth]{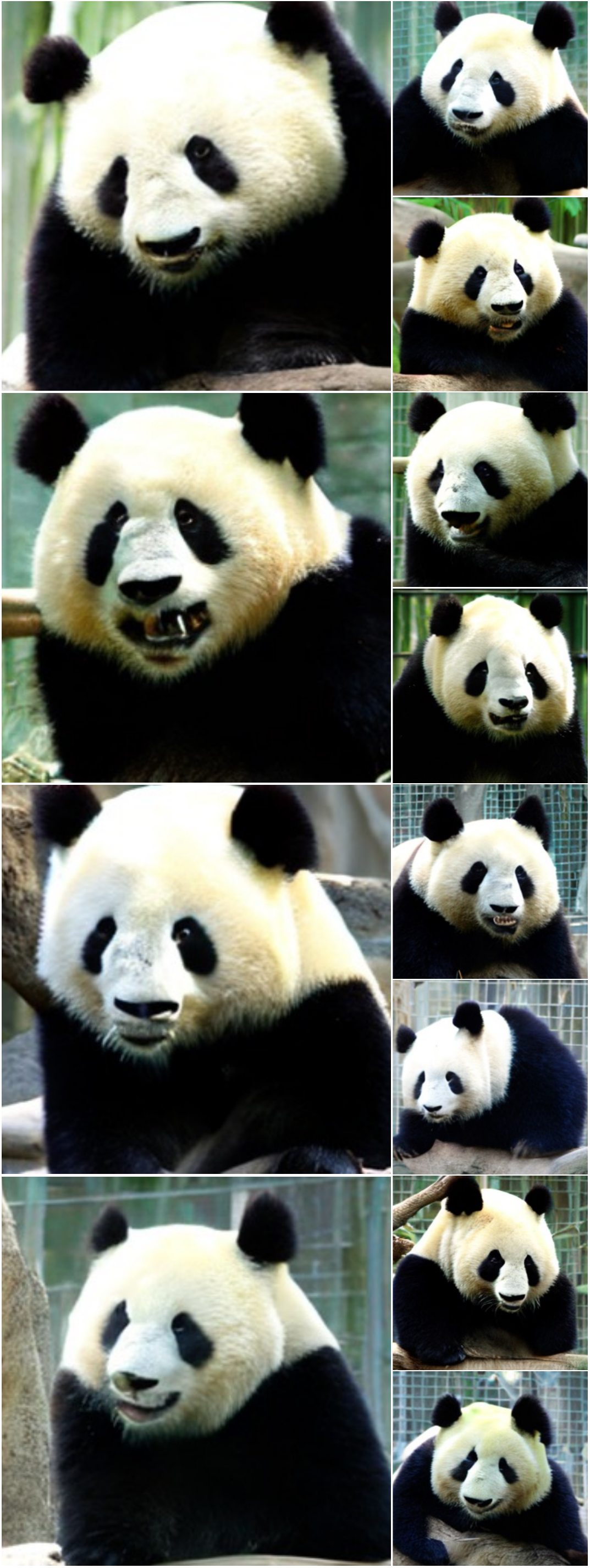}
  \caption{Images of n02510455 generated by FINE.}
  \label{fig:app1}
\end{figure}

\begin{figure}[t]
  \centering
  \includegraphics[width=0.9\linewidth]{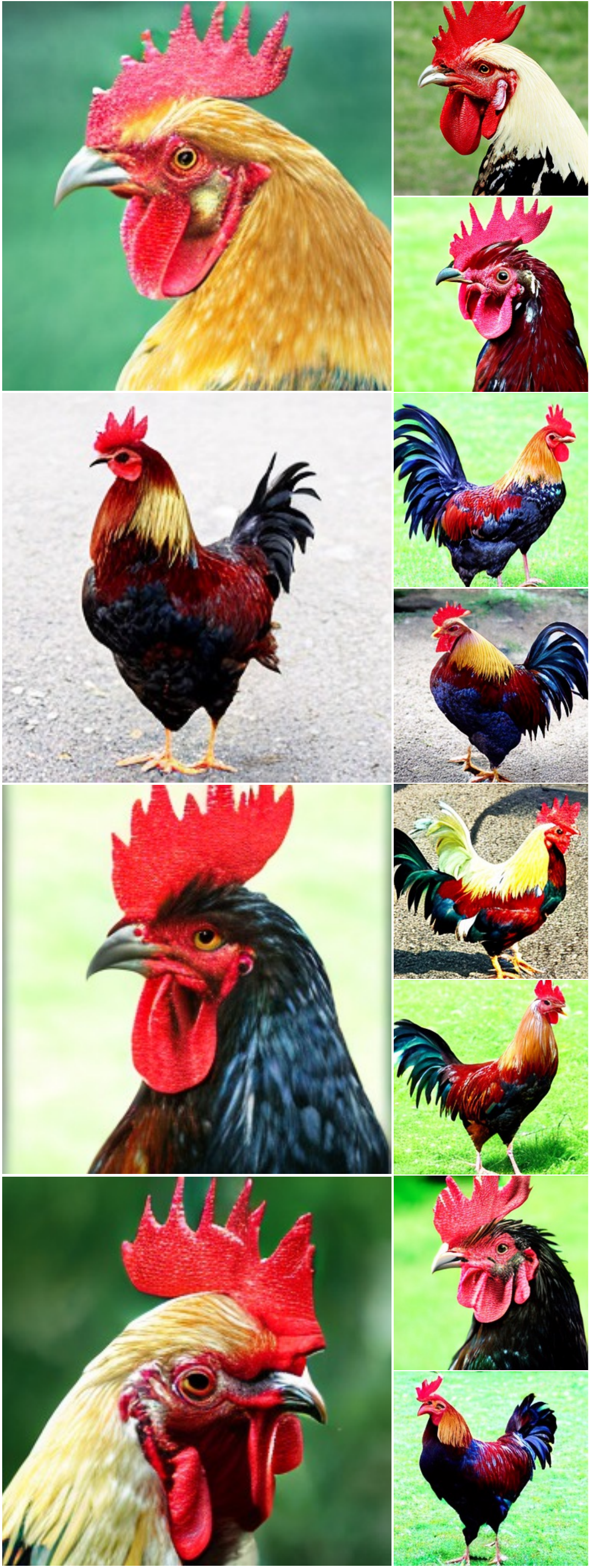}
  \caption{Images of n01514668 generated by FINE.}
  \label{fig:app2}
\end{figure}

\begin{figure}[t]
  \centering
  \includegraphics[width=0.9\linewidth]{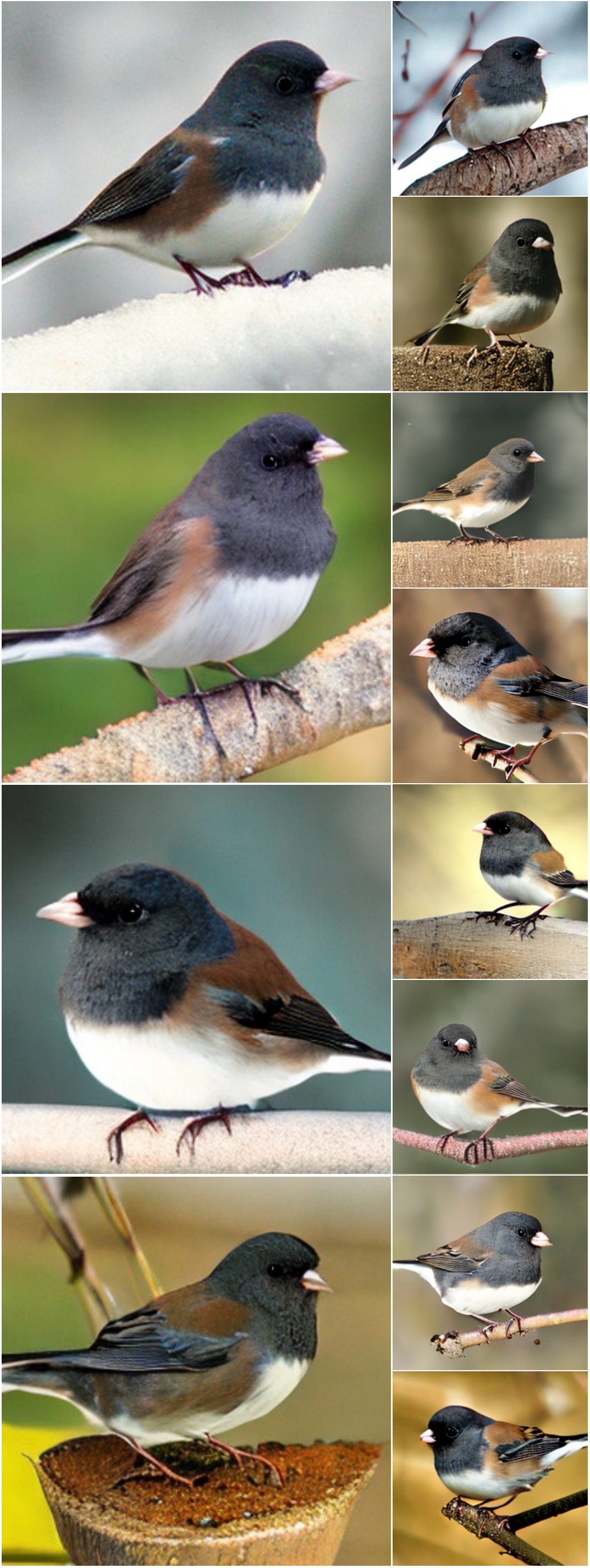}
  \caption{Images of n01534433 generated by FINE.}
  \label{fig:app3}
\end{figure}

\begin{figure}[t]
  \centering
  \includegraphics[width=0.9\linewidth]{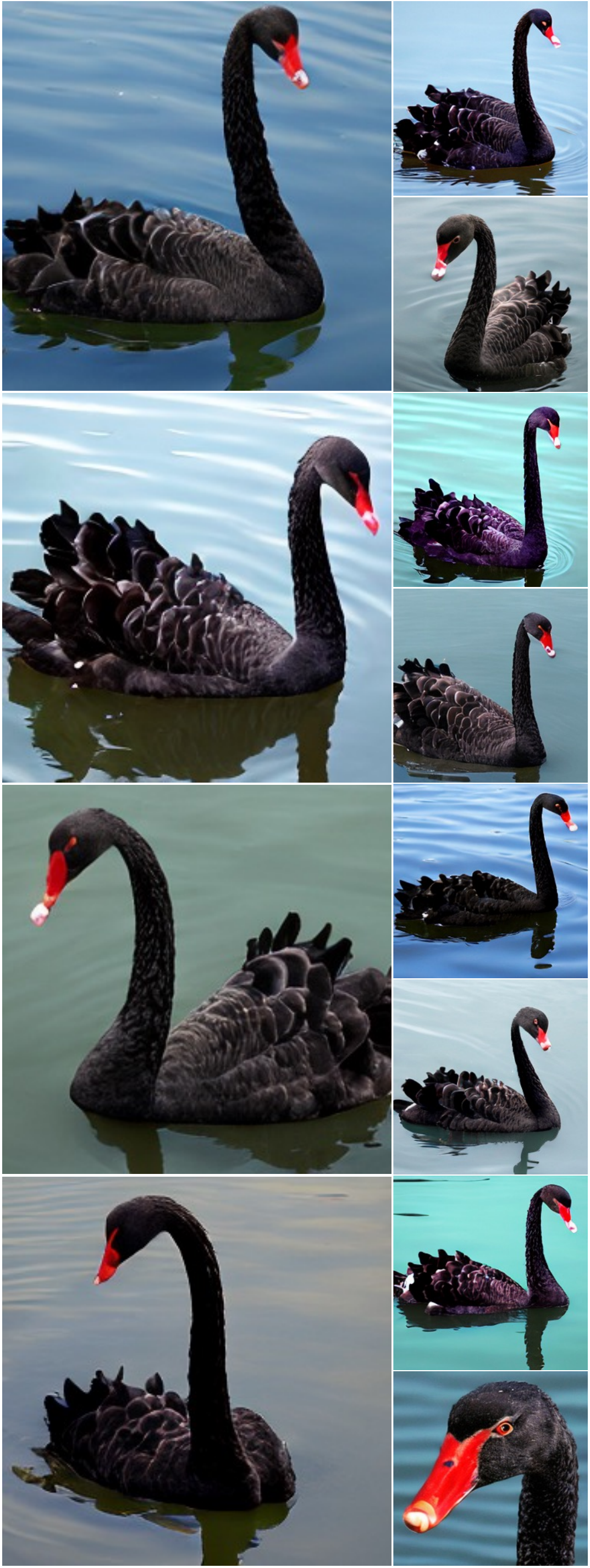}
  \caption{Images of n01860187 generated by FINE.}
  \label{fig:app4}
\end{figure}

\begin{figure}[t]
  \centering
  \includegraphics[width=0.9\linewidth]{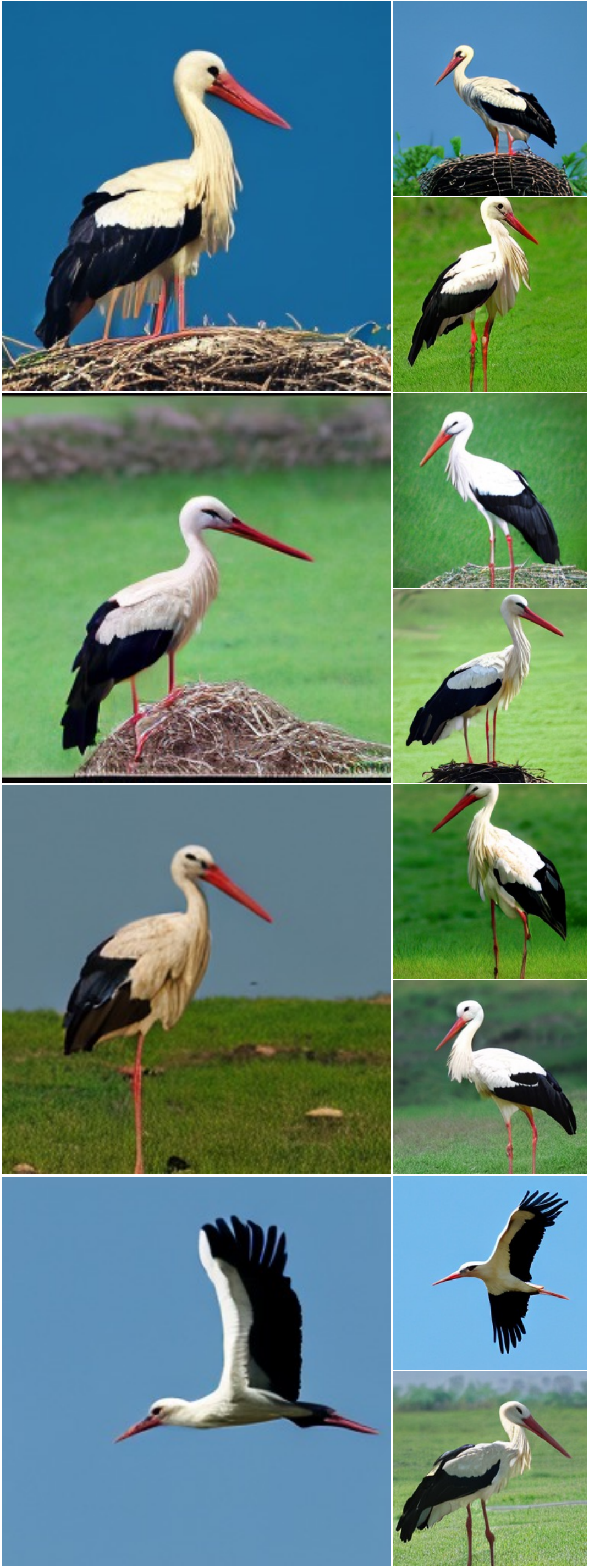}
  \caption{Images of n02002556 generated by FINE.}
  \label{fig:app5}
\end{figure}

\begin{figure}[t]
  \centering
  \includegraphics[width=0.9\linewidth]{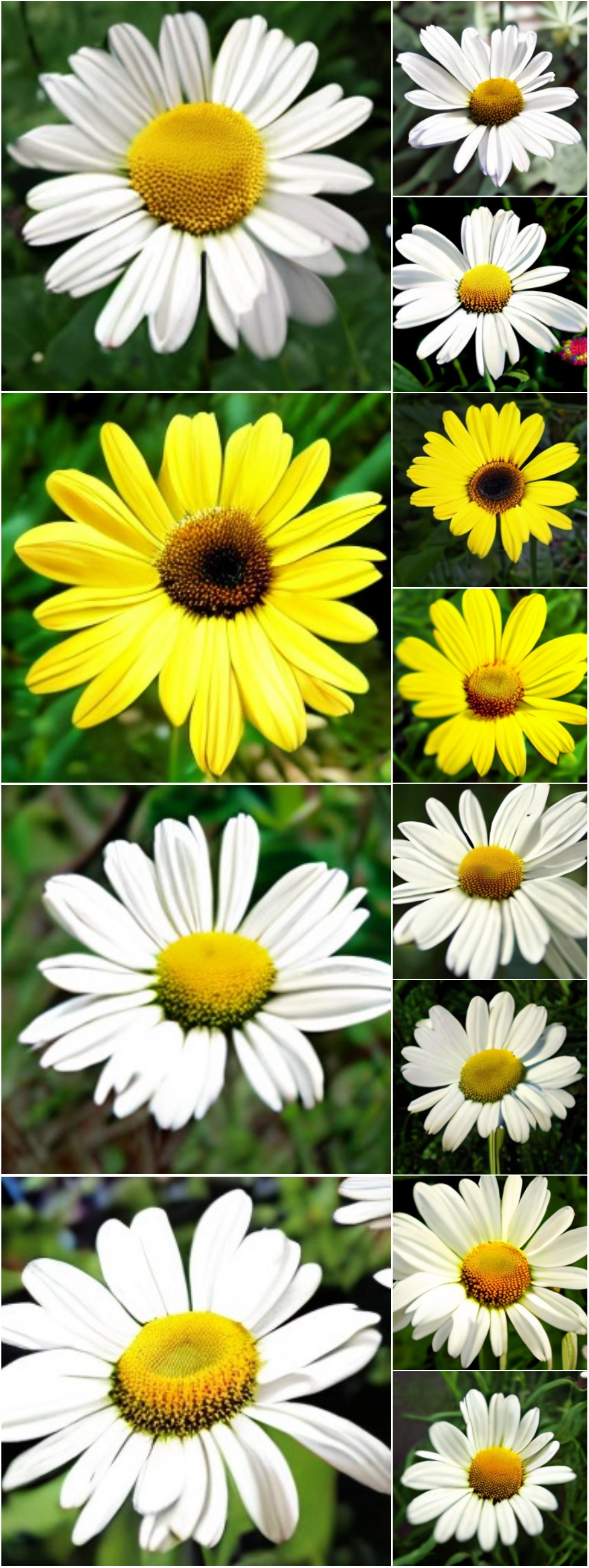}
  \caption{Images of n11939491 generated by FINE.}
  \label{fig:app6}
\end{figure}

\begin{figure}[t]
  \centering
  \includegraphics[width=0.9\linewidth]{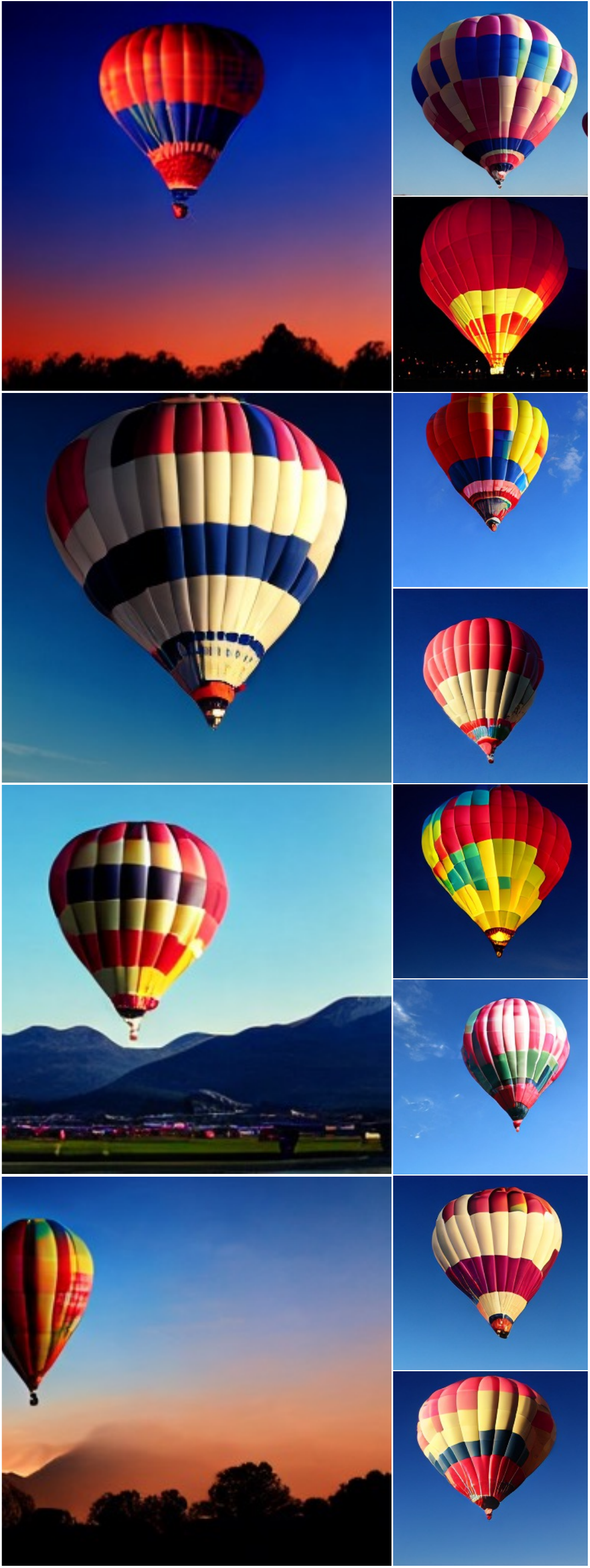}
  \caption{Images of n02782093 generated by FINE.}
  \label{fig:app7}
\end{figure}

\begin{figure}[t]
  \centering
  \includegraphics[width=0.9\linewidth]{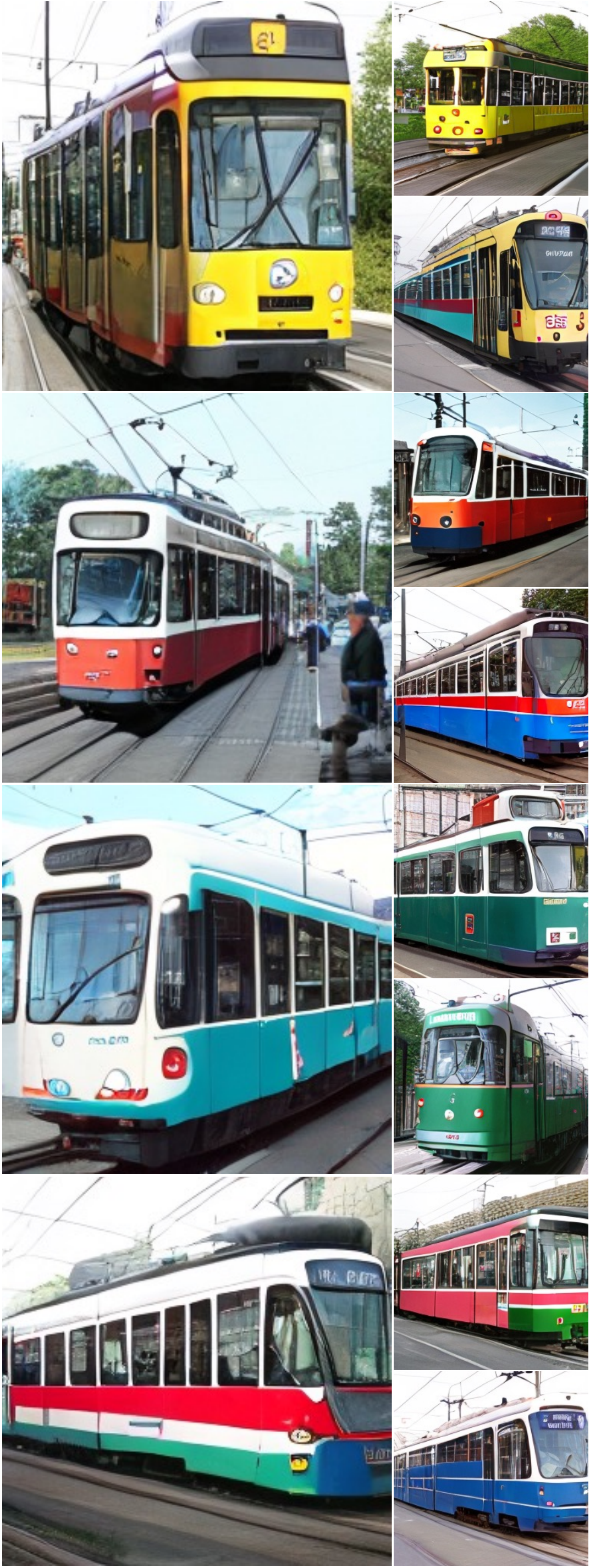}
  \caption{Images of n04335435 generated by FINE.}
  \label{fig:app8}
\end{figure}

We provide additional images generated by the DiT-L/2 model, initialized with FINE at a resolution of 256$\times$256, as illustrated in Figure~\ref{fig:app1}-\ref{fig:app8}.

\end{document}